%% file: acl_latex.tex
\pdfoutput=1
\documentclass[11pt]{article}
\usepackage[preprint]{acl}
\usepackage{times}
\usepackage{latexsym}
\usepackage[T1]{fontenc}
\usepackage[utf8]{inputenc}
\usepackage{microtype}
\usepackage{inconsolata}
\usepackage{graphicx}
\usepackage{booktabs}
\usepackage{amsmath}
\usepackage{xcolor}
\usepackage{float}
\usepackage{enumitem}
\usepackage{multirow}
\usepackage{tcolorbox}
\usepackage{setspace}

\NewDocumentCommand{\qingyun}
{ mO{} }{\textcolor{cyan}{\textsuperscript{\textit{Qingyun}}\textsf{\textbf{\small[#1]}}}}

\newcommand*{\inlinelargeimage}[1]{\raisebox{-0.15\baselineskip}
    {$\,$\includegraphics[height=0.9\baselineskip]{#1}$\,\,$}}
\newcommand{\emojim}{\inlinelargeimage{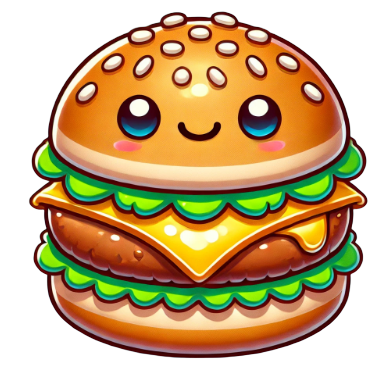}}

\title{\emojim MAC-Tuning: LLM Multi-Compositional Problem Reasoning with Enhanced Knowledge Boundary Awareness}

\author{
    \textbf{Junsheng Huang\textsuperscript{1,2}},
    \textbf{Zhitao He\textsuperscript{1}},
    \textbf{Yuchen Huang\textsuperscript{1}},
    \textbf{Sandeep Polisetty\textsuperscript{3}},
    \textbf{Qingyun Wang\textsuperscript{4}} \\
    \textbf{Yi R. (May) Fung\textsuperscript{1}}
    \\
    \textsuperscript{1}Hong Kong University of Science and Technology \\
    \textsuperscript{2}University of Illinois Urbana-Champaign 
    \hspace{1em}
    \textsuperscript{3}UMass Amherst \hspace{1em}\textsuperscript{4}William \& Mary
    \\
    jh103@illinois.edu ~~~~ yrfung@ust.hk
}

\begin{document}
\maketitle

\input{latex/0_abstract}
\input{latex/1_introduction}
\input{latex/2_method}
\input{latex/3_experiment}

\input{latex/4_conclusion}

\input{latex/5_limitation}

\bibliography{anthology,custom}


\appendix

\input{latex/appendix}

\end{document}

%% file: latex/0_abstract.tex
\begin{abstract}
The hallucination of non-existent facts by LLMs is an important problem given its widespread adoption across various applications. Previous research addresses this problem by analyzing the internal parameterized knowledge boundaries to estimate confidence. However, these studies focus on the single-problem setting and have not explored the more challenging multi-problem setting, which requires accurately answering multiple questions simultaneously. We introduce a novel method for the multi-problem setting, \textbf{M}ultiple \textbf{A}nswers and \textbf{C}onfidence Stepwise \textbf{Tuning} (\textbf{MAC-Tuning}), that separates the learning of answer prediction and confidence estimation during fine-tuning on instruction data. Extensive experiments demonstrate that our method outperforms baselines by up to 25\% in average precision.\footnote{We release our code and resource at \href{https://github.com/no-touch-fish/Multi-QA-Tuning}{MAC-Tuning}.}
\end{abstract}

%% file: latex/1_introduction.tex
\section{Introduction}
Large language models (LLMs) are widely used in knowledge-intensive scenarios, such as question answering \citep{gu-etal-2023-dont}, information retrieval \citep{ren2023tometwostageapproachmodelbased}, and recommendation systems \citep{Liu2023AFL}. Yet, they often produce non-existing facts when faced with questions outside their parametric knowledge, which undermines their reliability \citep{maynez-etal-2020-faithfulness,he2025mmboundaryadvancingmllmknowledge}. Many efforts have been dedicated to mitigating LLM hallucination, such as leveraging knowledge boundaries to constrain the reasoning scope of LLMs to help them better distinguish between reliable and unreliable information \cite{chen2024teachinglargelanguagemodels,liang-etal-2024-l, zhang-etal-2024-r,jin2024rwku}. Notably, these work mainly focus on the \textbf{single-problem setting}, where users repeatedly input questions and context for models to answer one by one. 
\begin{figure}[!t] 
    \centering
    \includegraphics[width=\columnwidth]{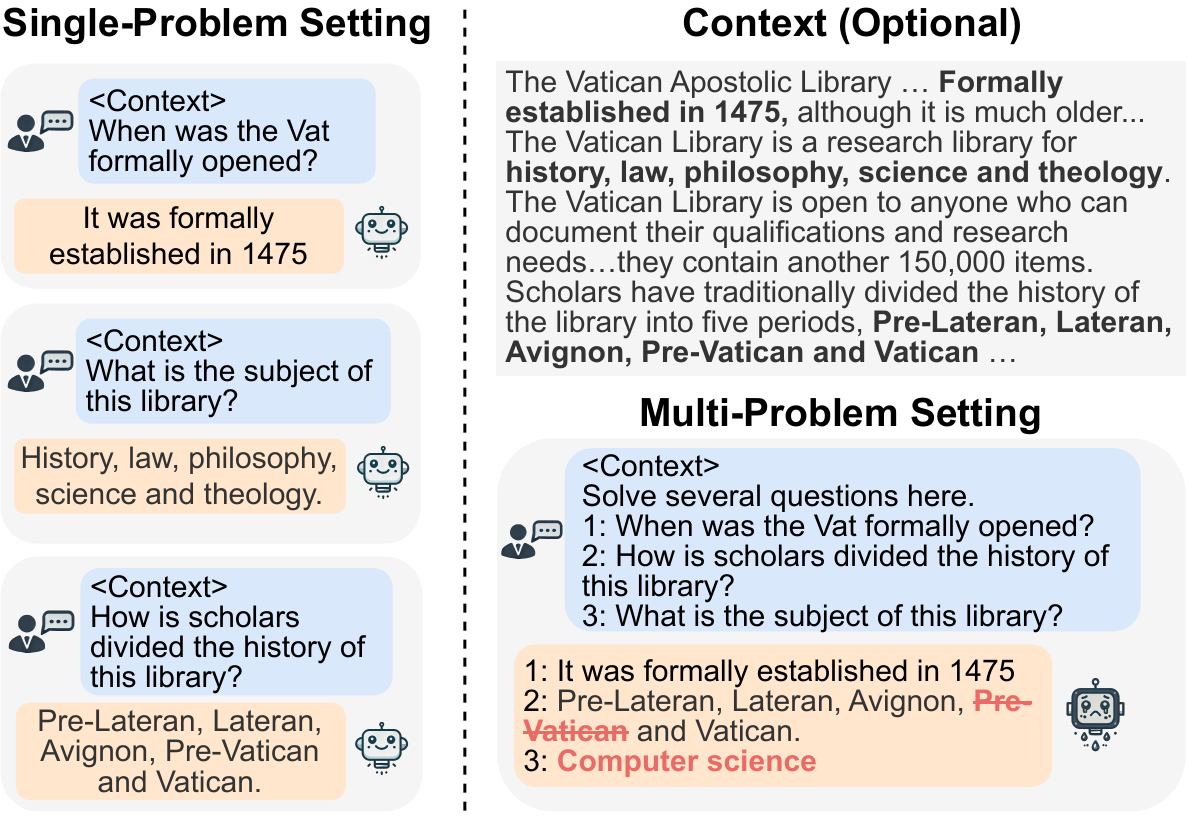} 
    \caption{An illustration of the multi-problem setting. \textcolor{red}{\textit{Red}} indicates that the LLM's output is inaccurate.} 
    \label{fig:Multi-Problem setting}
    \vspace{-1.2em}
\end{figure}

LLM hallucination in the \textbf{multi-problem setting} — in which a single input contains multiple distinct sub-questions with optional context for the model to extract and address — remains relatively underexplored. As seen in Figure \ref{fig:Multi-Problem setting}, this is a fundamentally challenging setting because the model must distinguish each sub-question, reason over different knowledge, and synthesize results cohesively. 
Undesirable overshadowing of context from one sub-question with another, and propagation of reasoning confusion, may compromise the reliability of LLMs in multi-problem answering (\citealp{cheng-etal-2023-batch}, \citealp{wang2024exploringzeroshotcapabilitiesllms}, \citealp{son2024multitaskinferencelargelanguage,li2024mosaicitfreecompositionaldata,he2025matpbenchmllmgoodautomated}). 
As LLM-based multi-problem reasoning becomes increasingly widespread due to its efficiency benefits in scenarios involving extensive shared contexts (e.g., task instructions, exemplars), reduced model access, and lower API costs, enhancing model confidence estimation calibration for this emerging class of reasoning demands growing attention and effort as well.

In this paper, we investigate the hallucinations in LLMs within the multi-problem setting and propose leveraging the knowledge boundary to simultaneously handle the composition of multiple problems. Inspired by \citet{zhang-etal-2024-r}, which advocates for encouraging the LLM to express confidence to reduce hallucinations, we introduce \textbf{M}ultiple \textbf{A}nswers and \textbf{C}onfidence Stepwise \textbf{Tuning} (\textbf{MAC-Tuning}) under the multi-problem setting. Our approach involves several key steps. First, we identify the knowledge boundary between parametric knowledge and the multi-problem dataset to extract uncertain questions. Next, we automatically label the model’s confidence for both certain and uncertain data. These labeled data are then used to create multiple question-answer data and multiple QA-Confidence data so we can train the original model by separating the learning process of ground-truth answers and confidence, which enhances performance and reliability.

Our contributions can be summarized as follows:
\begin{itemize}[leftmargin=10pt,topsep=2pt]
    \setlength\itemsep{0em}
    \item We are the first to explore LLM confidence estimation under the more challenging multi-problem setting, where LLMs must handle multiple problems simultaneously. \\[-1.6em]
    \item We propose MAC-Tuning, which separates the learning process of answer and confidence predictions for enhancing knowledge boundary awareness and reducing hallucination. \\[-1.6em]
    \item Through extensive experiments with different base models of varying sizes and various datasets, MAC-Tuning achieves an AP score gain of up to 25\% over baselines in LLM multi-problem reasoning. Finally, we share our insights discovered to motivate future work.
\end{itemize}

%% file: latex/2_method.tex
\section{Methodology}
Figure \ref{fig:pipeline} shows the data construction process for \textbf{M}ultiple \textbf{A}nswers and \textbf{C}onfidence \textbf{S}tepwise \textbf{T}uning (\textbf{MAC-Tuning}).
\vspace{-0.2em}

\subsection{Multi-Problem Tuning Data Construction}
\begin{figure}[ht]
    \centering
    \includegraphics[width=\columnwidth]{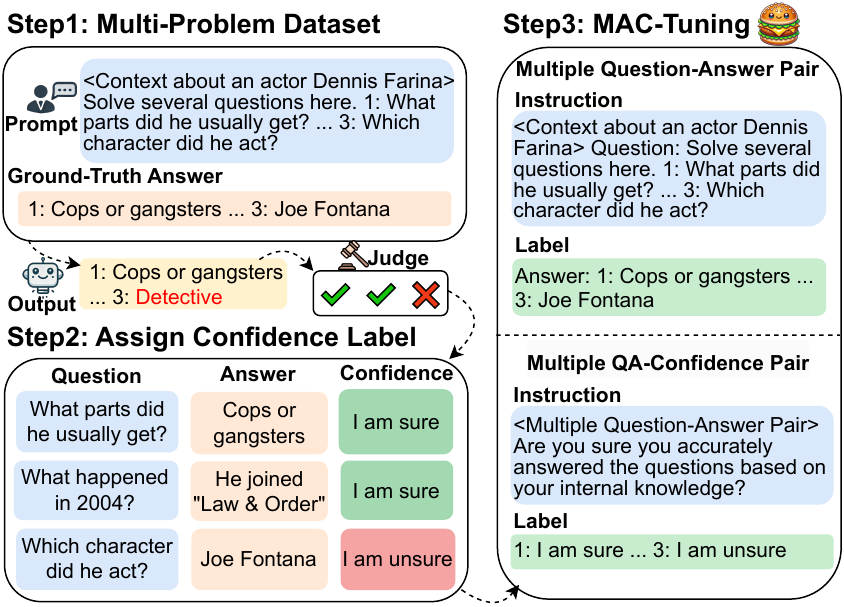}
    \caption{We first construct the Multi-Problem dataset, and then use it to generate Multi-Problem Tuning data.}
    \label{fig:pipeline}
\end{figure}

\vspace{-0.2em}
First, we combine \textbf{\textit{n}} single problems from original datasets to construct our initial Multi-Problem dataset. We utilize this to compare LLMs' outputs with ground-truth answers, for distinguishing the knowledge boundary between LLM parameters and instruction data. Specifically, for each individual problem in the multi-problem pair, we assign: ``\textit{I am sure}'' if the output aligns with ground-truth answer;  ``\textit{I am unsure}'' elsewise (e.g., Step 2 in Figure \ref{fig:pipeline}). With the assigned confidence labels, we construct Multi-Problem Tuning data as follows: 
\vspace{0.4em}
\noindent\textbf{Multiple QA pair $D_{MultQA}$}: We directly combine the questions and answers together, with \textit{Question} $q_i$ as input and \textit{Answer} $a_i$ as output label, to form $D_{MultQA} = [(q_{1},a_{1})...(q_{i},a_{i})...(q_{n},a_{n})]$. 

\noindent\textbf{Multiple QA-Confidence pair $D_{MultQA,C}$}: The input consists of an instruction for the LLM to express its confidence ({\em i.e,} certainty in correctness) for a given question-answer pair, while the output is the confidence level in linguistic form\footnote{The template is in Appendix \ref{appendix: template for QA-Confidence pair}}. 

\subsection{Training and Inference}
Using the Multi-Problem Tuning data, we conduct a two-step supervised fine-tuning process to train the model to answer questions and express confidence in a multi-problem setting. The objective for the first step, in answering question, is:
\vspace{-0.7em}
\begin{align}
\small
&\max_{\Theta_0} \sum_{(Q,A) \in D_{MultQA}} \log P(A | Q; \Theta_0)  
\end{align}
The objective for the second step, in expressing confidence, is:
\vspace{-0.5em}
\begin{align}
\small
&\max_{\Theta_1} \sum_{(Q,A,C) \in D_{MultQA,C}} \log P(C | Q, A; \Theta_1) 
\end{align}
where $Q$, $A$, and $C$ represent the sets of multiple questions, multiple answers, and multiple confidence levels, respectively. \(\Theta_0\) and \(\Theta_1\) represent the parameters of the base model and the model after the first step of fine-tuning, respectively.


%% file: latex/3_experiment.tex
\begin{table*}[ht]
\small
\resizebox{\textwidth}{!}{%
\begin{tabular}{c|cccccccc|cccc}
\toprule
\multirow{3}{*}{\textbf{Model}} &  
  \multicolumn{8}{c|}{\textbf{Independent}} &
  \multicolumn{4}{c}{\textbf{Sequential}} \\ \cline{2-13} 
 &
  \multicolumn{2}{c}{\textbf{CoQA}} &
  \multicolumn{2}{c}{\textbf{ParaRel}} &
  \multicolumn{2}{c}{\textbf{GSM}} &
  \multicolumn{2}{c|}{\textbf{MMLU}} &
  \multicolumn{2}{c}{\textbf{MTI-Bench}} &
  \multicolumn{2}{c}{\textbf{SQA}} \\ \cline{2-13} 
  & AP   & ECE  & AP   & ECE  & AP   & ECE  & AP   & ECE  & AP       & ECE      & AP      & ECE     \\ \midrule
LLaMA3    & 54.6 & 22.6 & 45.1 & 40.8 & 79.3 & 52.8 & 50.3 & 43.8 & 37.4     & 17.7     & 44.9    & 35.4    \\
QA-Only   & 66.3 & 15.1 & 53.7 & 12.6 & 75.3 & 36.1 & 58.5 & 17.9 & 45.0     & 16.9     & 56.6    & 21.0    \\
Single-QA & 65.5 & 28.9 & 73.5 & 10.7 & 56.6 & 44.5 & 58.3 & 25.7 & N/A      & N/A      & N/A     & N/A     \\
Merge-AC  & 67.4 & 17.0 & 73.0 & 65.3 & 75.1 & 44.8 & 58.5 & 18.3 & 38.3     & 33.7     & 49.2    & 31.7    \\
MAC-Tuning &
  \textbf{69.8} &
  \textbf{7.33} &
  \textbf{76.1} &
  \textbf{3.61} &
  \textbf{79.9} &
  \textbf{3.16} &
  \textbf{63.1} &
  \textbf{12.5} &
  \textbf{64.0} &
  \textbf{13.4} &
  \textbf{65.0} &
  \textbf{14.6} \\ \bottomrule
\end{tabular}%
}
\caption{This is the confidence calibration result (\%). We use one-shot CoT for GSM results. \textbf{Bold} font highlights the best performance for the dataset across different methods. We don't apply Single-QA to the \textit{Sequential} setting dataset, as doing so would disrupt the logical connections among the questions.}
\label{tab:main results}
\end{table*}
\vspace{-0.4em}
\section{Experiment}
\vspace{-0.2em}
\subsection{Dataset}
We validate the effectiveness of our method across different problem settings and datasets: for the \textit{Independent} setting, where the questions are not related to each other, we use the \textbf{CoQA} \citep{reddy-etal-2019-coqa}, \textbf{GSM} \citep{cobbe2021gsm8k}, \textbf{MMLU} \citep{hendryckstest2021}, and \textbf{ParaRel} \citep{Elazar2021MeasuringAI} datasets; for the \textit{Sequential} setting, where the questions are logically related to each other, we use the \textbf{MTI-Bench} \citep{son2024multitaskinferencelargelanguage} and \textbf{SQA} \citep{iyyer-etal-2017-search} datasets. These datasets are either Question Answer (QA) or Multiple Choice (MC) formats. Table \ref{tab:training stats} shows the details of the dataset. Further information on the distribution of certain and uncertain data among the training set across different datasets is detailed in Appendix~\ref{appendix:Dataset}. 
\begin{table}[htb]
\resizebox{\columnwidth}{!}{%
\begin{tabular}{@{}c|cccc|cc@{}}
\toprule
\multirow{2}{*}{\textbf{}} & \multicolumn{4}{c|}{\textbf{Independent}} & \multicolumn{2}{c}{\textbf{Sequential}} \\ 
\cline{2-7} 
& \textbf{CoQA} & \textbf{ParaRel} & \textbf{GSM} & \textbf{MMLU} & \textbf{MTI-Bench} & \textbf{SQA} \\ \midrule
\textbf{Train}    & 5006          & 7500             & 7468         & 2448          & 2400               & 3985         \\ 
\textbf{Test}     & 5011          & 5584             & 1319         & 2439          & 600                & 925          \\ 
\textbf{Type}     & QA           & QA                & QA           & MC            & QA                  & QA        \\ 
\bottomrule
\end{tabular}%
}
\caption{Statistics of the datasets.}
\label{tab:training stats}
\end{table}
\vspace{-1.5em}

\subsection{Evaluation Metrics}
\vspace{-0.2em}
We directly compare the LLM generation to the ground-truth answer for the Question-Answer format. For Multiple-Choice format, we check the choice (A, B, C, D) and the option in the LLM generation. Across both types of answer generation tasks, we consider three evaluation metrics: (1) \textbf{Average Precision (AP)}: We use AP to measure the precision in identifying and ranking relevant predictions. A higher AP score means the model has high certainty about correct answers and high uncertainty about wrong answers.
(2) \textbf{Expected Calibrated Error (ECE)}: We use ECE to measure how closely the predicted certainty reflects the true certainty of LLM \cite{chen-etal-2023-close}. Low ECE indicates better-calibrated predictions. (3) \textbf{Accuracy}: We compute accuracy as the fraction of correct responses amongst questions in which LLMs expressed certainty towards their answers. 

\subsection{Baselines}
We compare MAC-Tuning with the base model and its variants in the multi-problem settings. We use LLaMA3-8B-Instruct (\textbf{LLaMA3}) \citep{dubey2024llama3herdmodels} as the backbone. For baseline \textbf{QA-Only}, we fine-tune the base model directly using the Multiple Question-Answer pairs to evaluate the effectiveness of the traditional instruction tuning method under the multi-problem setting. For baseline \textbf{Single-QA}, we use single-problem data to fine-tune and directly apply it to the multi-problem setting. For baseline \textbf{Merge-AC}, instead of separating the learning process of ground-truth answers and confidence, we directly let the model learn multiple answers along with their corresponding confidence levels\footnote{Baseline examples are in Appendix \ref{appendix: Baseline methods}. Implementation details are in Appendix \ref{appendix: Implementation}.}. 

\subsection{Overall Performance}

In Table \ref{tab:main results}, we report the results on multi-problem setting from three single questions combined together. 
MAC-Tuning achieves the best AP score across all datasets, showing up to a 15\% improvement, along with a lower ECE. This suggests that after MAC-Tuning, the model becomes more adept at distinguishing between certain and uncertain questions, delivering more reliable results through improved confidence estimation in answer prediction. 
We also evaluate each model's accuracy on every dataset. MAC-Tuning consistently outperforms the base model in accuracy by up to 45.8\% and, on average, 23.7\%. The reason is that we separate the tasks of learning correct answers and confidence within a multi-problem setting. After learning the ground-truth answer, the LLM can better understand confidence, while still retaining its ability to extract information, respond accurately, and address multiple problems simultaneously.

\paragraph{Ablation on Different Component} 
We further test three variants of the MAC-Tuning method in the multi-problem setting: \textbf{QA-Only}, which is MAC-Tuning without the confidence component; \textbf{Single-QA}, where we evaluate MAC-Tuning with single problem data; and \textbf{Merge-AC}, where we evaluate MAC-Tuning without separating the learning process of ground-truth answers and confidence. As seen from the results in Table \ref{tab:main results}, \textbf{MAC-Tuning} has up to 25\% and, on average, 11\% AP improvement compared with \textbf{Merge-AC}, reflecting that separating the learning process of ground-truth answers and confidence is crucial in multi-problem setting, as LLM cannot learn both in one time. The performance of \textbf{Single-QA} is better than the base model but worse than \textbf{QA-Only} in most cases, showing that LLM can aware the knowledge boundary under single-problem setting and transfer it to multi-problem setting, but it is not sufficient for LLM to answer multiple problems simultaneously. 

\subsection{Investigation on Out of Domain Settings}
We perform MAC-Tuning on base model with \textit{Sequential} setting dataset SQA and test it on other datasets, with the results as presented in Table \ref{tab:OOD study}. Even on out-of-domain datasets, MAC-Tuning still outperforms the base model, showing that it can effectively learn the multi-problem setting and generalize across different domains.

\begin{table}[!h]
\small\centering
\begin{tabular}{@{}c|cccc@{}}
\toprule
\textbf{Metric}   & \textbf{CoQA} & \textbf{Pararel} & \textbf{MMLU} & \textbf{MTI-Bench}  \\ \midrule
Accuracy & 59.3 & 70.3    & 52.6 & 57.8 \\
AP score & 62.2 & 58.7    & 53.8 & 81.7 \\
ECE      & 10.4 & 9.64    & 8.95 & 16.1 \\ \bottomrule
\end{tabular}%
\caption{The result (\%) for MAC-Tuning on SQA dataset and test on other datasets.}
\label{tab:OOD study}
\end{table}
\vspace{-1.2em}

\subsection{Analysis on Various Number of Questions} \label{section: Combine Different Number of Problems}
\begin{figure}[t]
    \centering
    \includegraphics[width=0.9\linewidth]{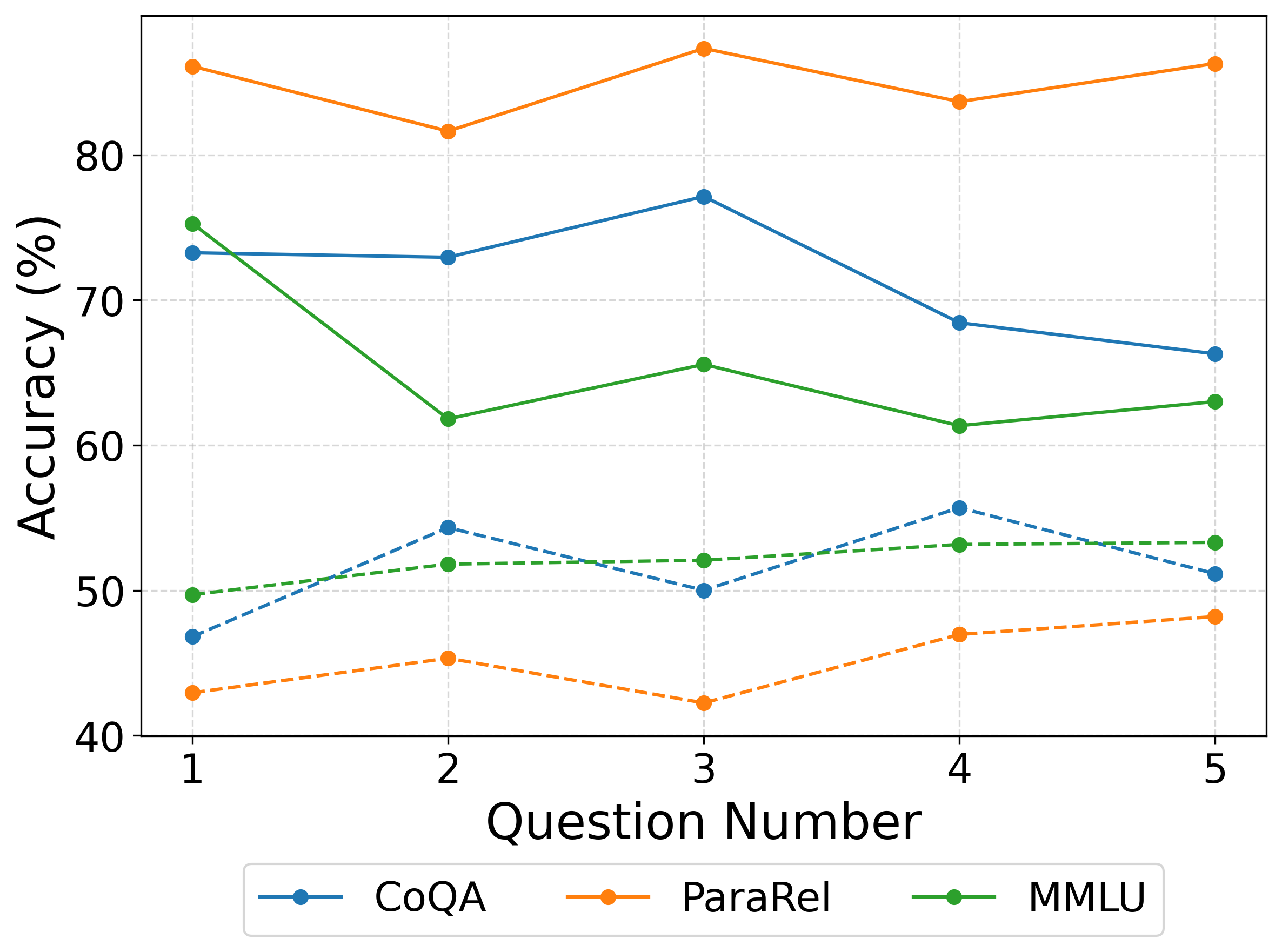}
    \caption{Accuracy for combining different number (\textbf{\textit{n}}) of single problem together. Solid lines represent MAC-Tuning, while dashed lines represent LLaMA3.}
    \label{fig:combine}
\end{figure}

We explore different numbers of questions in the multi-problem setting to investigate how this varies the accuracy. We only do this for three \textit{Independent} setting datasets, and the results are reported in Figure \ref{fig:combine}. MAC-Tuning consistently outperforms the base model in accuracy by at least 10.0\% and, on average, 26.8\%. For easy tasks like ParaRel, the ability of the base model to handle multiple problems simultaneously is even higher when compared with the traditional single-problem setting, indicating that LLM could leverage in-context learning and focus on relevant knowledge better under the multi-problem setting. 
However, for other datasets like MMLU, MAC-Tuning performs slightly worse as the question number increases. A reasonable explanation is that it is out of the base model's ability to learn too many hard tasks together but within effective scope to learn several easy tasks at the same time. The result for extremely large numbers of questions is in Appendix \ref{appendix: Extremely Large Multi-problem Scenarios}.

For future work, we believe it is meaningful to further explore whether mixing questions of varying difficulty and diversity in multi-problem settings leads to better scaling behavior \cite{qin2025scalinglawssyntheticdata}. This direction may help uncover strategies for enhanced model generalization. 

\subsection{Cross Task Transfer Study}
We fine-tune the model with question number \textit{\textbf{n}} = 3, and subsequently evaluate its performance on both single problem inputs (\textit{\textbf{n}} = 1) and more complex instances involving a higher number of questions (\textit{\textbf{n}} = 5). The evaluation on \textit{\textbf{n}} = 1 aims to examine whether the model retains its ability to accurately solve individual problems after being exposed to multi-problem setting training. Conversely, the evaluation on \textit{\textbf{n}} = 5 serves to assess the generation capability of MAC-Tuning when scaling to larger compositions beyond the training scope. This allows us to understand both the robustness and scalability of our proposed method across different levels of compositional complexity. The results are reported in Table \ref{tab: transfer}.
\begin{table}[ht]
\resizebox{\columnwidth}{!}{%
\begin{tabular}{@{}c|cccc@{}}
\toprule
\textbf{Question Number} & \textbf{CoQA} & \textbf{ParaRel} & \textbf{GSM} & \textbf{MMLU} \\ \midrule
n = 1                    & 78.8         & 84.2            &  71.1       & 54.6         \\
n = 5                    & 79.1         & 86.2            &  67.7       & 63.7         \\ \bottomrule
\end{tabular}%
}
\caption{Accuracy (\%) for MAC-Tuning with question number \textit{\textbf{n}} = 3 transferring to question number \textit{\textbf{n}} = 1 and question number \textit{\textbf{n}} = 5. We use one-shot CoT for GSM results.}
\label{tab: transfer}
\end{table}

From the result of single question inputs, we observe that accuracy increases on easier dataset (e.g. CoQA) but decreases on more challenging dataset like GSM, when compared to models fine-tuned specifically under that setting. This pattern indicates that LLM acquires underlying knowledge during MAC-Tuning rather than merely memorizing the patterns for multi-problem setting. In contrast, when fine-tuned on five-question inputs (\textit{\textbf{n}} = 5), the model's performance is comparable or even exceeds that of the baseline fine-tuned directly on \textit{\textbf{n}} = 5. These findings strengthen the statement we make in Section \ref{section: Combine Different Number of Problems}: while LLMs can efficiently learn multiple easy tasks, they exhibit difficulty when faced with several difficult tasks simultaneously.

\subsection{Analysis on Different Base Model}
\vspace{-0.3em}
Table \ref{tab:Qwen results} shows the result from changing the base model to Qwen2-7B-Instruct (\citealp{yang2024qwen2technicalreport}). We observe that the performance trends remain consistent even with a different base model. MAC-Tuning continues to demonstrate an average precision (AP) gain of up to near 24\% with a lower ECE, showcasing the effectiveness of learning ground-truth answers and confidence separately. 
\begin{table}[htb]
\resizebox{\columnwidth}{!}{%
\begin{tabular}{@{}c|cccc|cccc@{}}
\toprule
\textbf{} & \multicolumn{4}{c|}{\textbf{~Independent~}} & \multicolumn{4}{c}{\textbf{~Sequential~}} \\ 
\cline{2-9} 
\textbf{Approach} & \multicolumn{2}{c}{\textbf{\underline{~~ParaRel~~}}} & \multicolumn{2}{c|}{\textbf{\underline{~~~MMLU~~~~}}} & \multicolumn{2}{c}{\textbf{\underline{MTI-Bench}}} & \multicolumn{2}{c}{\textbf{\underline{~~~~~~~SQA~~~~~~}}} \\ 
          & AP       & ECE      & AP       & ECE      & AP       & ECE      & AP      & ECE     \\ \midrule
Vanilla    & 54.3     & 37.8     & 68.1     & 25.3     & 48.8     & 31.3     & 30.3    & 54.6    \\
MAC-Tuning & \textbf{78.7}     & \textbf{9.59}    & \textbf{73.0}    & \textbf{17.1}   & \textbf{53.3}      & \textbf{18.6}     & \textbf{47.7}   & \textbf{29.2}  \\ \bottomrule
\end{tabular}%
}
\caption{Confidence calibration result (\%) for Qwen2-7B-Instruct, with \textbf{bold} denoting the top performance.}
\label{tab:Qwen results}
\end{table}
\vspace{-1.4em}

\subsection{Analysis on Different Model Size}
We compare base models of different sizes to study how model size affects performance and the confidence calibration results for Llama-3.2-3B (\citealp{dubey2024llama3herdmodels}) and Phi-3.5-mini-Instruct (\citealp{abdin2024phi3technicalreporthighly}) is shown in Table \ref{tab:llama3.2-3B-results} and Table \ref{tab:phi3.5-mini-results} respectively. Despite using different base models of varying sizes, the results indicate consistent performance patterns. For smaller models, the accuracy improvement after \textbf{MAC-Tuning} is more evident, indicating enhanced ability to differentiate between certain and uncertain questions.

\begin{table}[htb]
\resizebox{\columnwidth}{!}{%
\begin{tabular}{@{}c|cccc|cccc@{}}
\toprule
\textbf{} & \multicolumn{4}{c|}{\textbf{~Independent~}} & \multicolumn{4}{c}{\textbf{~Sequential~}} \\ 
\cline{2-9} 
\textbf{Approach} & \multicolumn{2}{c}{\textbf{\underline{~~ParaRel~~}}} & \multicolumn{2}{c|}{\textbf{\underline{~~~CoQA~~~~}}} & \multicolumn{2}{c}{\textbf{\underline{MTI-Bench}}} & \multicolumn{2}{c}{\textbf{\underline{~~~~~~~SQA~~~~~~}}} \\ 
          & AP       & ECE      & AP       & ECE      & AP       & ECE      & AP      & ECE     \\ \midrule
Vanilla    & 30.7     & 70.3     & 46.0     & 45.0     & 34.3     & 70.3     & 35.6    & 45.0    \\
MAC-Tuning & \textbf{55.5}     & \textbf{33.5}    & \textbf{62.4}    & \textbf{33.5}   & \textbf{35.5}      & \textbf{33.5}     & \textbf{44.3}   & \textbf{33.5}  \\ \bottomrule
\end{tabular}%
}
\caption{Confidence calibration result (\%) for Llama-3.2-3B, with \textbf{bold} denoting the top performance across different methods.}
\label{tab:llama3.2-3B-results}
\end{table}

\begin{table}[htb]
\resizebox{\columnwidth}{!}{%
\begin{tabular}{@{}c|cccc|cccc@{}}
\toprule
\textbf{} & \multicolumn{4}{c|}{\textbf{~Independent~}} & \multicolumn{4}{c}{\textbf{~Sequential~}} \\ 
\cline{2-9} 
\textbf{Approach} & \multicolumn{2}{c}{\textbf{\underline{~~ParaRel~~}}} & \multicolumn{2}{c|}{\textbf{\underline{~~~CoQA~~~~}}} & \multicolumn{2}{c}{\textbf{\underline{MTI-Bench}}} & \multicolumn{2}{c}{\textbf{\underline{~~~~~~~SQA~~~~~~}}} \\ 
          & AP       & ECE      & AP       & ECE      & AP       & ECE      & AP      & ECE     \\ \midrule
Vanilla    & 58.0     & 22.8     & 56.0     & 32.9     & 21.4     & 29.1     & 96.6    & 33.7    \\
MAC-Tuning & \textbf{70.2}     & \textbf{14.2}    & \textbf{68.2}    & \textbf{29.0}   & \textbf{68.7}      & \textbf{22.6}     & \textbf{52.3}   & \textbf{23.9}  \\ \bottomrule
\end{tabular}%
}
\caption{Confidence calibration result (\%) for Phi-3.5-mini-Instruct, with \textbf{bold} denoting the top performance across different methods.}
\label{tab:phi3.5-mini-results}
\end{table}

\subsection{Human Evaluation}
We randomly selected and evaluated 100 examples from the ParaRel dataset. The human annotator was shown only the query and the ground-truth answer, and asked to assess the factual correctness of each model’s output without knowing the confidence label (“I am sure” or “I am unsure”). We then compared the accuracy rates between responses that the model labeled as “I am sure” versus those labeled as “I am unsure.” For answers the model labeled as “I am sure”: Human evaluation confirmed a factual accuracy of \textbf{89.2\%}. For answers the model labeled as “I am unsure”: The human-verified factual accuracy was only \textbf{41.2\%}.

These results provide strong empirical evidence for our central claim. The substantial 48-percentage-point gap demonstrates that the confidence learned by MAC-Tuning is not merely an artifact of automatic metrics. Instead, it reflects a genuine, human-perceptible distinction in answer quality. This strong alignment with human judgment validates the reliability and real-world applicability of our method.

%% file: latex/4_conclusion.tex
\section{Conclusion}
\vspace{-0.3em}
In this paper, we introduce a novel method, MAC-Tuning, to enhance large language model (LLM) confidence calibration and reasoning robustness in the challenging yet underexplored multi-problem scenario. 
Our proposed approach automatically constructs multi-problem setting question-answer pairs with confidence annotations for identifying the intrinsic knowledge gap between parametric knowledge and instructional data. With this data constructed, we guide the LLM to better reason on answer prediction and confidence estimation separately, in multi-problem setting. 
Extensive experiments across different datasets show that our method significantly improves performance in areas where the original LLM struggles. 

%% file: latex/5_limitation.tex
\section*{Limitation}
While our work provides valuable insight on the new Multiple Question setting and introduces an innovative fine-tuning method, there are several limitations to acknowledge. First, although we experimented with various prompts, as is typical in prompt-based LLM studies, we cannot ensure that slight changes in prompts would not significantly alter the results. Second, due to constraints of cost, time, and computational resources, we selected a subset of experiments that we believe to be informative and representative. However, additional experiments across a wider range of datasets and LLMs might provide further insights. Lastly, in this new setting, there may be other underlying reasons for the experimental results. Future work will aim to address these limitations by expanding datasets and conducting new experiments to explore other potential factors affecting performance.

%% file: latex/appendix.tex
\section{Appendix}\label{sec:appendix}

\subsection{Full Case for Examples of Introduction} \label{appendix: full case for exmaples of introduction}
\begin{figure}[htb] 
    \centering
    \includegraphics[width=\columnwidth]{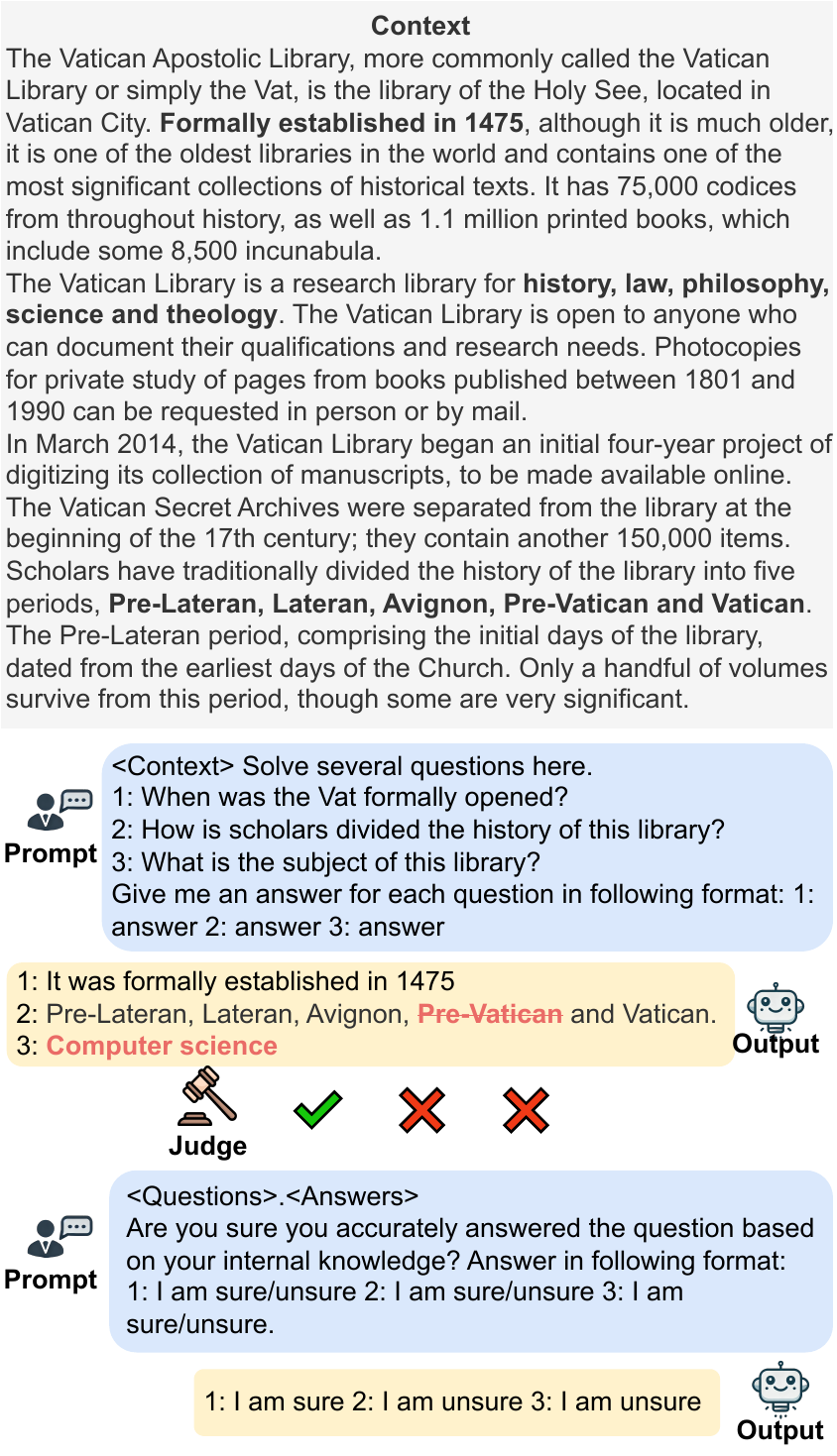} 
    \caption{The full case of examples in introduction in Multiple Problem setting. Red context indicates that LLM's output is inaccurate. The second answer lacks the information of "Pre-Vatican" and the third answer contains a completely factual error. After MAC-Tuning, LLM show uncertainty towards answering this two previously incorrect questions.}
    \label{fig:full case for exmaples of introduction}
\end{figure}

Full case for the examples in introduction can be found in Figure \ref{fig:full case for exmaples of introduction}.

\subsection{Related Work}
\noindent\textbf{Hallucination:} Large language models (LLMs) are widely used in knowledge-intensive scenarios, such as question answering \citep{gu-etal-2023-dont,he2025mmboundaryadvancingmllmknowledge}, information retrieval \citep{ren2023tometwostageapproachmodelbased,huang2025cultureclipempoweringclipcultural} and recommendation systems \citep{Liu2023AFL}. However, LLMs have tendency to generate non-existing facts when faced with questions that are out of their parametric knowledge \citep{maynez-etal-2020-faithfulness}. Many efforts are dedicated to mitigating hallucinations in LLMs, such as retrieval-augmented generation (\citealp{gao2024retrievalaugmentedgenerationlargelanguage},\citealp{peng2023check}), 
multi-agent debate (\citealp{he-etal-2023-lego,du2023improvingfactualityreasoninglanguage,he2024agentscourtbuildingjudicialdecisionmaking,sun2023principledrivenselfalignmentlanguagemodels,he2025advancinglanguagemultiagentlearning}), and 
model confidence calibration (\citealp{zhang-etal-2024-r,hu-etal-2023-wont,he2025mmboundaryadvancingmllmknowledge}). 

\noindent\textbf{Knowledge Boundary:} There are many different ways to utilize knowledge boundary to reduce LLM hallucination. \citet{liang-etal-2024-learning}'s work uses merged knowledge probing and consistency checking methods to help LLM express their internal knowledge. \citet{chen2024teachinglargelanguagemodels}'s work leverages LLM internal signals to let LLM know their unknowns. \citet{zhang-etal-2024-r} utilize knowledge boundary to instruct LLM say "I don't know". It is a popular way to use confidence to express knowledge boundary of LLMs and we also follow this.

\noindent\textbf{Multiple Problem Setting:} Current LLM research has predominantly focused on single problem setting. There are only a few works focusing on this new setting. \citet{cheng-etal-2023-batch} propose batch prompting that prompts LLMs with single independent problems batched together following few-shot exemplars together. \citet{son2024multitaskinferencelargelanguage} goes further by researching sequential datasets and develops the first multi-task benchmark (MTI-Bench). \citet{wang2024exploringzeroshotcapabilitiesllms} pays attention to zero-shot cases of multi-problem setting and  design a new benchmark ZeMPEB. \citet{li2024mosaicitfreecompositionaldata} analyze different strategy under independent setting, where single questions are combined into various constraint formats without sharing context between them.
Despite these efforts, the multi-problem setting presents significant challenges. For example, \citet{wang2024exploringzeroshotcapabilitiesllms} shows that in zero-shot setting, LLMs consistently perform worse when selecting indices of texts for a given class label with multiple mixed-source reasoning problems. Similarly, for few-shot setting, \citet{cheng2023batchpromptingefficientinference} and \citet{lin2024batchpromptaccomplish} have found that the overall accuracy decreases with the increase in batch size.
Notably, this setting is also meaningful in real-world applications: for independent scenario, batching unrelated queries can reduce model calls and API costs; for sequential scenario, where questions share context—such as in math problem solving, data processing, or software debugging—the correctness of each intermediate reasoning step is critical. Overall, hallucination and performance instability under the multi-problem setting are still under-explored and present significant challenges for current LLMs.

\subsection{Template for QA-Confidence pair} \label{appendix: template for QA-Confidence pair}
\begin{tcolorbox}[colback=gray!10, colframe=black, width=\columnwidth, sharp corners]
\small{Question: \textit{<Question>}. Answer: \textit{<Answer>}. Are you sure you accurately answered the question based on your internal knowledge?

1: \textit{<Confidence>} 2: \textit{<Confidence>} 3: \textit{<Confidence>}}
\end{tcolorbox}

\subsection{Dataset Details} \label{appendix:Dataset}
We carry out our experiments across six datasets, described as follows.
\begin{itemize} 
    \item \textbf{GSM} \citep{cobbe2021gsm8k}: a dataset containing high-quality grade school math problems created by the OpenAI group. These problems require between 2 and 8 steps to solve, primarily involving a sequence of elementary calculations with basic arithmetic operations such as addition, subtraction, multiplication, and division to arrive at the final answer. We directly use 7.5k training data and 1k testing data in our Question Answer setting. 
    \item \textbf{Pararel} \citep{Elazar2021MeasuringAI}: a dataset containing factual knowledge with a variety of prompts and relationships, originally created for mask prediction. In Question Answer setting, we employ the modified dataset from \citet{zhang-etal-2024-r}.
    \item \textbf{MMLU} \citep{hendryckstest2021}: a dataset covering different subjects and difficulty. It tests both world knowledge and problem solving ability, which has good granularity and breath. We directly use the modified dataset from \citet{zhang-etal-2024-r} in our Multiple Choice setting.
    \item \textbf{CoQA} \citep{reddy-etal-2019-coqa}: a dataset designed to evaluate the ability of models to understand and generate answers in a conversational setting. We randomly pick 5k training dataset from theirs. In Question Answer setting, we combine multiple questions together under the same "story" category in the dataset.
    \item \textbf{MTI Bench} \citep{son2024multitaskinferencelargelanguage}: a comprehensive evaluation benchmark encompassing 5,000 instances across 25 tasks. We pick the sequential part of this benchmark and divide it into 800 training data and 200 test data.
    \item \textbf{SQA} \citep{iyyer-etal-2017-search}: a dataset designed to explore the task of answering sequences of inter-related questions on HTML tables. We pick 5 sequential questions for each HTML table and have 3985 training data.
\end{itemize}

\subsection{Formula and Calculation Details} \label{appendix: Formula and Calculation details}

\noindent \textbf{Average Precision (AP) Score} measures the performance of a binary classifier's confidence rankings. It corresponds to the area under the Precision-Recall curve. It is calculated as follows: 
\[
\mathrm{AP} = \sum_{k=1}^{n} \Bigl(R_k - R_{k-1}\Bigr)\times P_k
\]
where \textit{k} is the number of data at current thread with precision $P_{k}$ and recall $R_{k}$. \textit{n} is the total data number. The confidence is the weighted average of certain prediction probability and uncertain prediction probability.

\noindent \textbf{Expected Calibrated Error (ECE)} indicates how well a model's predicted probabilities match the true likelihood of an event. We split the predictions into 10 bins based on the certain prediction probability, then compare the average predicted probability to the actual proportion of positive samples (correct cases) in each bin. It is calculated as follows:
\[
\mathrm{ECE} = \sum_{m=1}^{10} \frac{\lvert B_m \rvert}{n} \bigl\lvert \overline{p}_m - \overline{y}_m \bigr\rvert
\]
where \textit{m} is the bin number with corresponding average predicted probability $\overline{p}_m$ and actual proportion of positive samples $\overline{y}_m$.

\subsection{Implementation} \label{appendix: Implementation}
We use HuggingFace PEFT \citep{peft} to conduct LoRA fine-tuning \citep{hu2021loralowrankadaptationlarge}. We set the training epoch to 3, learning rate to $1\mathbf{e}^{-5}$, LoRa rank to 8, and LoRa scaling factor to 32. The batch size is 1 and the temperature is 0. All experiments are implemented on Nvidia A100-40GB GPUs.

\subsection{Case Study} \label{appendix:Case Study}
\begin{figure*}[ht] 
    \centering
    \includegraphics[width=\textwidth]{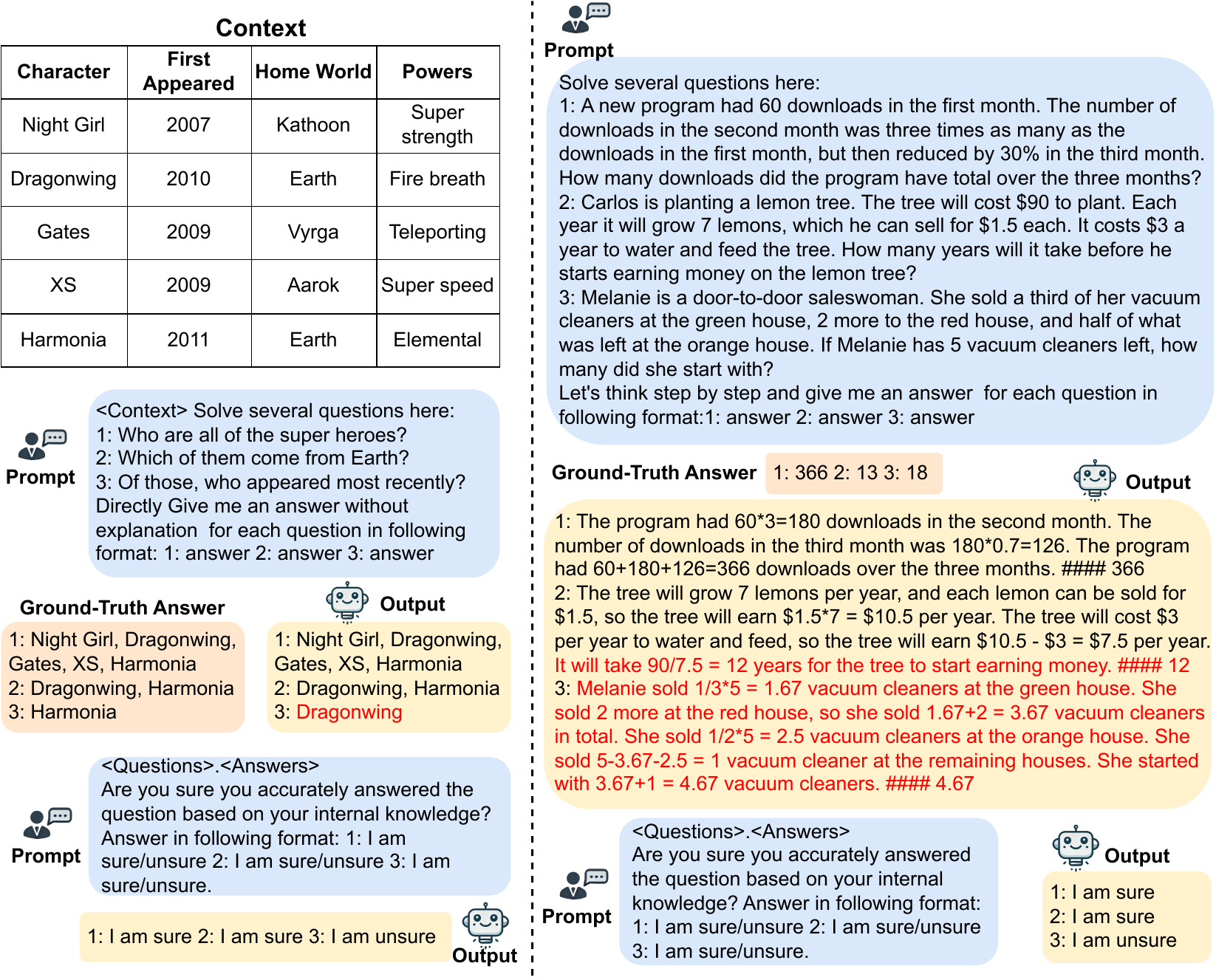}
    \caption{Specific cases for MAC-Tuning under the multiple problem setting with question number \textit{n} = 3. Red-highlighted context indicates inaccuracies in the LLM's output. The left example is drawn from the \textit{Sequential} setting dataset (SQA), while the right example is from the \textit{Independent} setting dataset (GSM), with one-shot context omitted for conciseness.}
    \label{fig:case study}
\end{figure*}

We show two specific cases for MAC-Tuning under the multiple problem setting with question number \textit{n} = 3 in Figure \ref{fig:case study}. The example on the left is from the SQA dataset, in which a table context is given and the LLM need to answer sequential questions based on the table. LLM answers correctly and shows certainty to first two questions, so these two questions will be counted into accuracy calculation. It answers wrong and shows uncertainty to the third question, which achieves the refusal behavior that we aim to see. The example on the right is from the GSM dataset. The LLM gives wrong answers to the second question but indicates certainty, which means this is a failure case.

\subsection{Detailed Information for Variant Methods} \label{appendix: Baseline methods}
\begin{figure*}[b] 
    \centering
    \includegraphics[width=\textwidth]{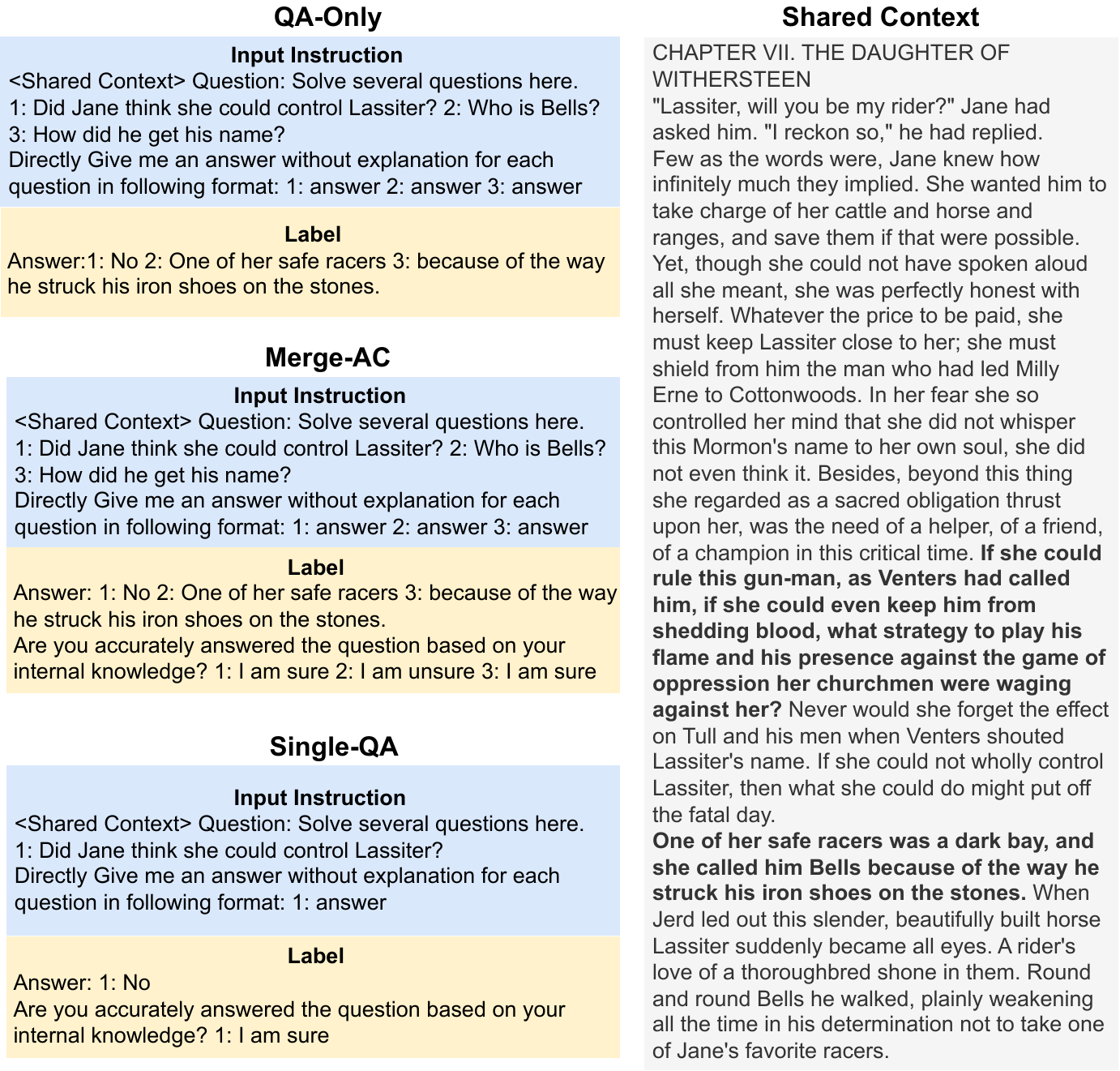} 
    \caption{A specific case to show how baseline methods are doing the fine-tuning. The answers are derived from the highlighted portions of the context. In QA-Only, the input is the Question instruction, and the output is the Answer. In Merge-AC, the output includes both the Answer and its Confidence. Single-QA is the single-problem variant of Merge-AC.}
    \label{fig:baseline methods}
\end{figure*}

The detailed example for different baseline methods is shown in Figure \ref{fig:baseline methods}. 

\subsection{Extremely Large Multi-problem Scenarios} \label{appendix: Extremely Large Multi-problem Scenarios}
We also conduct experiments to include larger values of question numbers (\textit{\textbf{n}} = 10 and 15). These results are shown in table \ref{tab:extreme large question number table}.
\begin{table}[ht]
\centering
\resizebox{\columnwidth}{!}{%
\begin{tabular}{@{}c|ccccccc@{}}
\toprule
\textbf{Pararel} & \textbf{n=1} & \textbf{n=2} & \textbf{n=3} & \textbf{n=4} & \textbf{n=5} & \textbf{n=10} & \textbf{n=15} \\ \midrule
LLaMA3 & 42.9 & 45.3 & 42.2 & 47.0 & 48.2 & 48.2 & 49.6 \\
MAC-Tuning & 86.1 & 81.6 & 88.0 & 83.7 & 86.3 & 82.6 & 77.1 \\ \bottomrule
\end{tabular}%
}
\caption{Accuracy (\%) comparison between LLaMA3 and MAC-Tuning on ParaRel dataset with extremely large question number of combined questions $n$.}
\label{tab:extreme large question number table}
\end{table}

The extended results show that MAC-Tuning consistently and significantly outperforms the Llama3 baseline across all tested values of n. Notably, even at n = 15, our method achieves 77.1\% accuracy, maintaining a substantial margin of over 27.5 percentage points against the baseline (49.6\%).

One notable point is that for different values of \textit{\textbf{n}}, we still use 3 training epochs for fairness. However, we only have around 5000 training questions for the Pararel dataset. Thus, for a large value of \textit{\textbf{n}} (taking \textit{\textbf{n = 10}} as an example), each epoch only has 500 training data and the model might be under training because of this.

Overall, it provides a clearer picture of our method’s performance scalability and demonstrates its robustness in handling more complex, large-scale multi-problem inputs.

\subsection{Certainty Distribution of the Training Dataset}
We demonstrate the certainty distribution of the training dataset under Multiple Problem setting with question number \textit{n} = 3 in Figure \ref{fig:training}:
\begin{figure*}
    \centering
    \includegraphics[width=\textwidth]{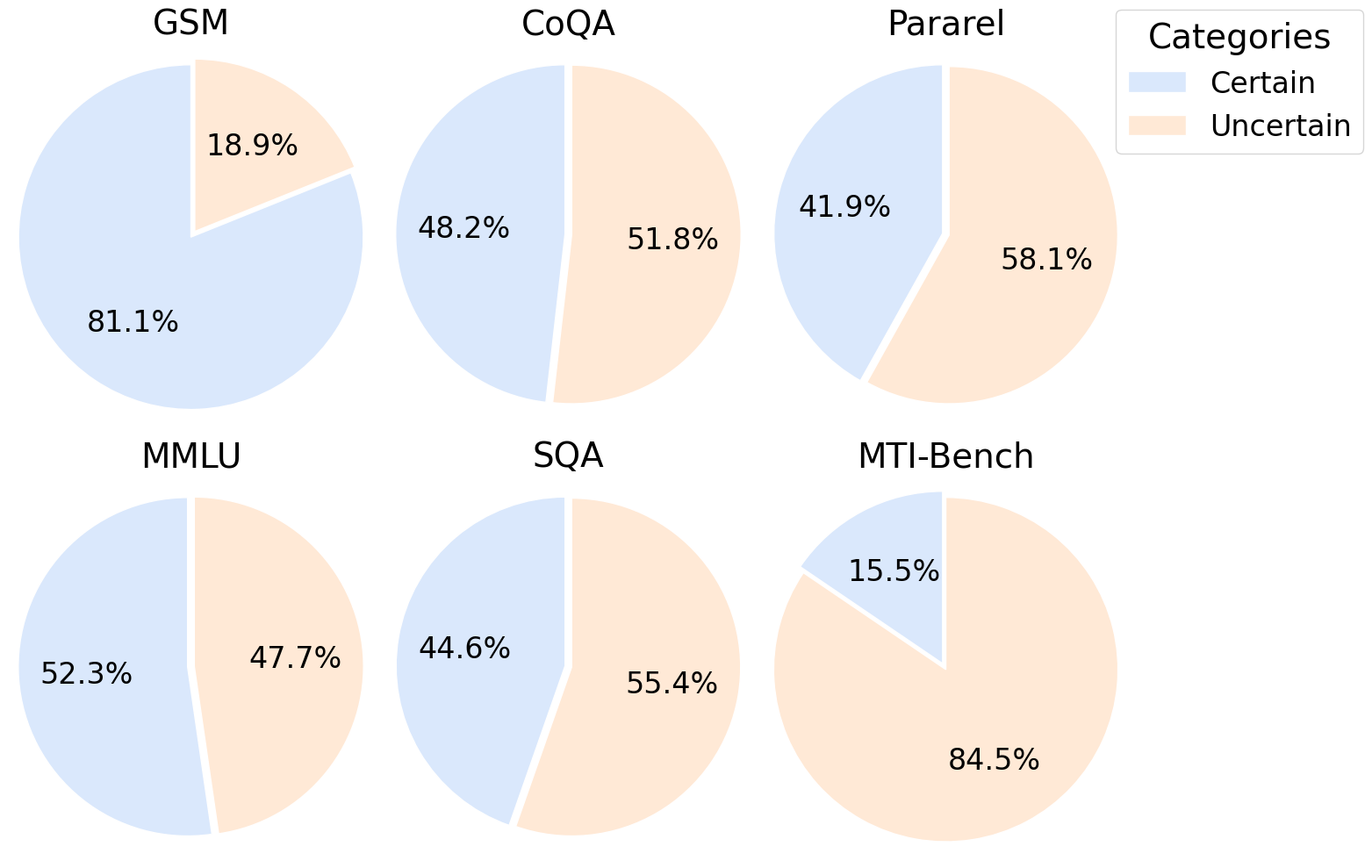}
    \caption{Certainty distribution of the training set under multi-problem setting with \textit{\textbf{n}} = 3}
    \label{fig:training}
\end{figure*}

%% file: acl_latex.bbl
\begin{thebibliography}{38}
\expandafter\ifx\csname natexlab\endcsname\relax\def\natexlab#1{#1}\fi

\bibitem[{Abdin et~al.(2024)Abdin, Aneja, Awadalla, Awadallah, Awan, Bach, Bahree, Bakhtiari, Bao, Behl, Benhaim, Bilenko, Bjorck, Bubeck, Cai, Cai, Chaudhary, Chen, Chen, Chen, Chen, Chen, Cheng, Chopra, Dai, Dixon, Eldan, Fragoso, Gao, Gao, Gao, Garg, Giorno, Goswami, Gunasekar, Haider, Hao, Hewett, Hu, Huynh, Iter, Jacobs, Javaheripi, Jin, Karampatziakis, Kauffmann, Khademi, Kim, Kim, Kurilenko, Lee, Lee, Li, Li, Liang, Liden, Lin, Lin, Liu, Liu, Liu, Liu, Liu, Luo, Madan, Mahmoudzadeh, Majercak, Mazzola, Mendes, Mitra, Modi, Nguyen, Norick, Patra, Perez-Becker, Portet, Pryzant, Qin, Radmilac, Ren, de~Rosa, Rosset, Roy, Ruwase, Saarikivi, Saied, Salim, Santacroce, Shah, Shang, Sharma, Shen, Shukla, Song, Tanaka, Tupini, Vaddamanu, Wang, Wang, Wang, Wang, Wang, Wang, Ward, Wen, Witte, Wu, Wu, Wyatt, Xiao, Xu, Xu, Xu, Xue, Yadav, Yang, Yang, Yang, Yang, Yu, Yuan, Zhang, Zhang, Zhang, Zhang, Zhang, Zhang, Zhang, and Zhou}]{abdin2024phi3technicalreporthighly}
Marah Abdin, Jyoti Aneja, Hany Awadalla, Ahmed Awadallah, Ammar~Ahmad Awan, Nguyen Bach, Amit Bahree, Arash Bakhtiari, Jianmin Bao, Harkirat Behl, Alon Benhaim, Misha Bilenko, Johan Bjorck, Sébastien Bubeck, Martin Cai, Qin Cai, Vishrav Chaudhary, Dong Chen, Dongdong Chen, Weizhu Chen, Yen-Chun Chen, Yi-Ling Chen, Hao Cheng, Parul Chopra, Xiyang Dai, Matthew Dixon, Ronen Eldan, Victor Fragoso, Jianfeng Gao, Mei Gao, Min Gao, Amit Garg, Allie~Del Giorno, Abhishek Goswami, Suriya Gunasekar, Emman Haider, Junheng Hao, Russell~J. Hewett, Wenxiang Hu, Jamie Huynh, Dan Iter, Sam~Ade Jacobs, Mojan Javaheripi, Xin Jin, Nikos Karampatziakis, Piero Kauffmann, Mahoud Khademi, Dongwoo Kim, Young~Jin Kim, Lev Kurilenko, James~R. Lee, Yin~Tat Lee, Yuanzhi Li, Yunsheng Li, Chen Liang, Lars Liden, Xihui Lin, Zeqi Lin, Ce~Liu, Liyuan Liu, Mengchen Liu, Weishung Liu, Xiaodong Liu, Chong Luo, Piyush Madan, Ali Mahmoudzadeh, David Majercak, Matt Mazzola, Caio César~Teodoro Mendes, Arindam Mitra, Hardik Modi, Anh Nguyen,
  Brandon Norick, Barun Patra, Daniel Perez-Becker, Thomas Portet, Reid Pryzant, Heyang Qin, Marko Radmilac, Liliang Ren, Gustavo de~Rosa, Corby Rosset, Sambudha Roy, Olatunji Ruwase, Olli Saarikivi, Amin Saied, Adil Salim, Michael Santacroce, Shital Shah, Ning Shang, Hiteshi Sharma, Yelong Shen, Swadheen Shukla, Xia Song, Masahiro Tanaka, Andrea Tupini, Praneetha Vaddamanu, Chunyu Wang, Guanhua Wang, Lijuan Wang, Shuohang Wang, Xin Wang, Yu~Wang, Rachel Ward, Wen Wen, Philipp Witte, Haiping Wu, Xiaoxia Wu, Michael Wyatt, Bin Xiao, Can Xu, Jiahang Xu, Weijian Xu, Jilong Xue, Sonali Yadav, Fan Yang, Jianwei Yang, Yifan Yang, Ziyi Yang, Donghan Yu, Lu~Yuan, Chenruidong Zhang, Cyril Zhang, Jianwen Zhang, Li~Lyna Zhang, Yi~Zhang, Yue Zhang, Yunan Zhang, and Xiren Zhou. 2024.
\newblock \href {http://arxiv.org/abs/2404.14219} {Phi-3 technical report: A highly capable language model locally on your phone}.

\bibitem[{Chen et~al.(2024)Chen, Liang, Wang, Liang, Xiao, Wei, Chen, Hao, Han, and Wang}]{chen2024teachinglargelanguagemodels}
Lida Chen, Zujie Liang, Xintao Wang, Jiaqing Liang, Yanghua Xiao, Feng Wei, Jinglei Chen, Zhenghong Hao, Bing Han, and Wei Wang. 2024.
\newblock \href {http://arxiv.org/abs/2406.10881} {Teaching large language models to express knowledge boundary from their own signals}.

\bibitem[{Chen et~al.(2023)Chen, Yuan, Cui, Liu, and Ji}]{chen-etal-2023-close}
Yangyi Chen, Lifan Yuan, Ganqu Cui, Zhiyuan Liu, and Heng Ji. 2023.
\newblock \href {https://doi.org/10.18653/v1/2023.acl-long.75} {A close look into the calibration of pre-trained language models}.
\newblock In \emph{Proceedings of the 61st Annual Meeting of the Association for Computational Linguistics (Volume 1: Long Papers)}, pages 1343--1367, Toronto, Canada. Association for Computational Linguistics.

\bibitem[{Cheng et~al.(2023{\natexlab{a}})Cheng, Kasai, and Yu}]{cheng-etal-2023-batch}
Zhoujun Cheng, Jungo Kasai, and Tao Yu. 2023{\natexlab{a}}.
\newblock \href {https://doi.org/10.18653/v1/2023.emnlp-industry.74} {Batch prompting: Efficient inference with large language model {API}s}.
\newblock In \emph{Proceedings of the 2023 Conference on Empirical Methods in Natural Language Processing: Industry Track}, pages 792--810, Singapore. Association for Computational Linguistics.

\bibitem[{Cheng et~al.(2023{\natexlab{b}})Cheng, Kasai, and Yu}]{cheng2023batchpromptingefficientinference}
Zhoujun Cheng, Jungo Kasai, and Tao Yu. 2023{\natexlab{b}}.
\newblock \href {http://arxiv.org/abs/2301.08721} {Batch prompting: Efficient inference with large language model apis}.

\bibitem[{Cobbe et~al.(2021)Cobbe, Kosaraju, Bavarian, Chen, Jun, Kaiser, Plappert, Tworek, Hilton, Nakano, Hesse, and Schulman}]{cobbe2021gsm8k}
Karl Cobbe, Vineet Kosaraju, Mohammad Bavarian, Mark Chen, Heewoo Jun, Lukasz Kaiser, Matthias Plappert, Jerry Tworek, Jacob Hilton, Reiichiro Nakano, Christopher Hesse, and John Schulman. 2021.
\newblock Training verifiers to solve math word problems.
\newblock \emph{arXiv preprint arXiv:2110.14168}.

\bibitem[{Du et~al.(2023)Du, Li, Torralba, Tenenbaum, and Mordatch}]{du2023improvingfactualityreasoninglanguage}
Yilun Du, Shuang Li, Antonio Torralba, Joshua~B. Tenenbaum, and Igor Mordatch. 2023.
\newblock \href {http://arxiv.org/abs/2305.14325} {Improving factuality and reasoning in language models through multiagent debate}.

\bibitem[{Dubey et~al.(2024)Dubey, Jauhri, Pandey, Kadian, Al-Dahle, Letman, Mathur, Schelten, Yang, Fan, Goyal, Hartshorn, Yang, Mitra, Sravankumar, Korenev, Hinsvark, Rao, Zhang, Rodriguez, Gregerson, Spataru, Roziere, Biron, Tang, Chern, Caucheteux, Nayak, Bi, Marra, McConnell, Keller, Touret, Wu, Wong, Ferrer, Nikolaidis, Allonsius, Song, Pintz, Livshits, Esiobu, Choudhary, Mahajan, Garcia-Olano, Perino, Hupkes, Lakomkin, AlBadawy, Lobanova, Dinan, Smith, Radenovic, Zhang, Synnaeve, Lee, Anderson, Nail, Mialon, Pang, Cucurell, Nguyen, Korevaar, Xu, Touvron, Zarov, Ibarra, Kloumann, Misra, Evtimov, Copet, Lee, Geffert, Vranes, Park, Mahadeokar, Shah, van~der Linde, Billock, Hong, Lee, Fu, Chi, Huang, Liu, Wang, Yu, Bitton, Spisak, Park, Rocca, Johnstun, Saxe, Jia, Alwala, Upasani, Plawiak, Li, Heafield, Stone, El-Arini, Iyer, Malik, Chiu, Bhalla, Rantala-Yeary, van~der Maaten, Chen, Tan, Jenkins, Martin, Madaan, Malo, Blecher, Landzaat, de~Oliveira, Muzzi, Pasupuleti, Singh, Paluri, Kardas, Oldham, Rita,
  Pavlova, Kambadur, Lewis, Si, Singh, Hassan, Goyal, Torabi, Bashlykov, Bogoychev, Chatterji, Duchenne, Çelebi, Alrassy, Zhang, Li, Vasic, Weng, Bhargava, Dubal, Krishnan, Koura, Xu, He, Dong, Srinivasan, Ganapathy, Calderer, Cabral, Stojnic, Raileanu, Girdhar, Patel, Sauvestre, Polidoro, Sumbaly, Taylor, Silva, Hou, Wang, Hosseini, Chennabasappa, Singh, Bell, Kim, Edunov, Nie, Narang, Raparthy, Shen, Wan, Bhosale, Zhang, Vandenhende, Batra, Whitman, Sootla, Collot, Gururangan, Borodinsky, Herman, Fowler, Sheasha, Georgiou, Scialom, Speckbacher, Mihaylov, Xiao, Karn, Goswami, Gupta, Ramanathan, Kerkez, Gonguet, Do, Vogeti, Petrovic, Chu, Xiong, Fu, Meers, Martinet, Wang, Tan, Xie, Jia, Wang, Goldschlag, Gaur, Babaei, Wen, Song, Zhang, Li, Mao, Coudert, Yan, Chen, Papakipos, Singh, Grattafiori, Jain, Kelsey, Shajnfeld, Gangidi, Victoria, Goldstand, Menon, Sharma, Boesenberg, Vaughan, Baevski, Feinstein, Kallet, Sangani, Yunus, Lupu, Alvarado, Caples, Gu, Ho, Poulton, Ryan, Ramchandani, Franco, Saraf,
  Chowdhury, Gabriel, Bharambe, Eisenman, Yazdan, James, Maurer, Leonhardi, Huang, Loyd, Paola, Paranjape, Liu, Wu, Ni, Hancock, Wasti, Spence, Stojkovic, Gamido, Montalvo, Parker, Burton, Mejia, Wang, Kim, Zhou, Hu, Chu, Cai, Tindal, Feichtenhofer, Civin, Beaty, Kreymer, Li, Wyatt, Adkins, Xu, Testuggine, David, Parikh, Liskovich, Foss, Wang, Le, Holland, Dowling, Jamil, Montgomery, Presani, Hahn, Wood, Brinkman, Arcaute, Dunbar, Smothers, Sun, Kreuk, Tian, Ozgenel, Caggioni, Guzmán, Kanayet, Seide, Florez, Schwarz, Badeer, Swee, Halpern, Thattai, Herman, Sizov, Guangyi, Zhang, Lakshminarayanan, Shojanazeri, Zou, Wang, Zha, Habeeb, Rudolph, Suk, Aspegren, Goldman, Damlaj, Molybog, Tufanov, Veliche, Gat, Weissman, Geboski, Kohli, Asher, Gaya, Marcus, Tang, Chan, Zhen, Reizenstein, Teboul, Zhong, Jin, Yang, Cummings, Carvill, Shepard, McPhie, Torres, Ginsburg, Wang, Wu, U, Saxena, Prasad, Khandelwal, Zand, Matosich, Veeraraghavan, Michelena, Li, Huang, Chawla, Lakhotia, Huang, Chen, Garg, A, Silva, Bell,
  Zhang, Guo, Yu, Moshkovich, Wehrstedt, Khabsa, Avalani, Bhatt, Tsimpoukelli, Mankus, Hasson, Lennie, Reso, Groshev, Naumov, Lathi, Keneally, Seltzer, Valko, Restrepo, Patel, Vyatskov, Samvelyan, Clark, Macey, Wang, Hermoso, Metanat, Rastegari, Bansal, Santhanam, Parks, White, Bawa, Singhal, Egebo, Usunier, Laptev, Dong, Zhang, Cheng, Chernoguz, Hart, Salpekar, Kalinli, Kent, Parekh, Saab, Balaji, Rittner, Bontrager, Roux, Dollar, Zvyagina, Ratanchandani, Yuvraj, Liang, Alao, Rodriguez, Ayub, Murthy, Nayani, Mitra, Li, Hogan, Battey, Wang, Maheswari, Howes, Rinott, Bondu, Datta, Chugh, Hunt, Dhillon, Sidorov, Pan, Verma, Yamamoto, Ramaswamy, Lindsay, Lindsay, Feng, Lin, Zha, Shankar, Zhang, Zhang, Wang, Agarwal, Sajuyigbe, Chintala, Max, Chen, Kehoe, Satterfield, Govindaprasad, Gupta, Cho, Virk, Subramanian, Choudhury, Goldman, Remez, Glaser, Best, Kohler, Robinson, Li, Zhang, Matthews, Chou, Shaked, Vontimitta, Ajayi, Montanez, Mohan, Kumar, Mangla, Albiero, Ionescu, Poenaru, Mihailescu, Ivanov, Li, Wang,
  Jiang, Bouaziz, Constable, Tang, Wang, Wu, Wang, Xia, Wu, Gao, Chen, Hu, Jia, Qi, Li, Zhang, Zhang, Adi, Nam, Yu, Wang, Hao, Qian, He, Rait, DeVito, Rosnbrick, Wen, Yang, and Zhao}]{dubey2024llama3herdmodels}
Abhimanyu Dubey, Abhinav Jauhri, Abhinav Pandey, Abhishek Kadian, Ahmad Al-Dahle, Aiesha Letman, Akhil Mathur, Alan Schelten, Amy Yang, Angela Fan, Anirudh Goyal, Anthony Hartshorn, Aobo Yang, Archi Mitra, Archie Sravankumar, Artem Korenev, Arthur Hinsvark, Arun Rao, Aston Zhang, Aurelien Rodriguez, Austen Gregerson, Ava Spataru, Baptiste Roziere, Bethany Biron, Binh Tang, Bobbie Chern, Charlotte Caucheteux, Chaya Nayak, Chloe Bi, Chris Marra, Chris McConnell, Christian Keller, Christophe Touret, Chunyang Wu, Corinne Wong, Cristian~Canton Ferrer, Cyrus Nikolaidis, Damien Allonsius, Daniel Song, Danielle Pintz, Danny Livshits, David Esiobu, Dhruv Choudhary, Dhruv Mahajan, Diego Garcia-Olano, Diego Perino, Dieuwke Hupkes, Egor Lakomkin, Ehab AlBadawy, Elina Lobanova, Emily Dinan, Eric~Michael Smith, Filip Radenovic, Frank Zhang, Gabriel Synnaeve, Gabrielle Lee, Georgia~Lewis Anderson, Graeme Nail, Gregoire Mialon, Guan Pang, Guillem Cucurell, Hailey Nguyen, Hannah Korevaar, Hu~Xu, Hugo Touvron, Iliyan Zarov,
  Imanol~Arrieta Ibarra, Isabel Kloumann, Ishan Misra, Ivan Evtimov, Jade Copet, Jaewon Lee, Jan Geffert, Jana Vranes, Jason Park, Jay Mahadeokar, Jeet Shah, Jelmer van~der Linde, Jennifer Billock, Jenny Hong, Jenya Lee, Jeremy Fu, Jianfeng Chi, Jianyu Huang, Jiawen Liu, Jie Wang, Jiecao Yu, Joanna Bitton, Joe Spisak, Jongsoo Park, Joseph Rocca, Joshua Johnstun, Joshua Saxe, Junteng Jia, Kalyan~Vasuden Alwala, Kartikeya Upasani, Kate Plawiak, Ke~Li, Kenneth Heafield, Kevin Stone, Khalid El-Arini, Krithika Iyer, Kshitiz Malik, Kuenley Chiu, Kunal Bhalla, Lauren Rantala-Yeary, Laurens van~der Maaten, Lawrence Chen, Liang Tan, Liz Jenkins, Louis Martin, Lovish Madaan, Lubo Malo, Lukas Blecher, Lukas Landzaat, Luke de~Oliveira, Madeline Muzzi, Mahesh Pasupuleti, Mannat Singh, Manohar Paluri, Marcin Kardas, Mathew Oldham, Mathieu Rita, Maya Pavlova, Melanie Kambadur, Mike Lewis, Min Si, Mitesh~Kumar Singh, Mona Hassan, Naman Goyal, Narjes Torabi, Nikolay Bashlykov, Nikolay Bogoychev, Niladri Chatterji, Olivier
  Duchenne, Onur Çelebi, Patrick Alrassy, Pengchuan Zhang, Pengwei Li, Petar Vasic, Peter Weng, Prajjwal Bhargava, Pratik Dubal, Praveen Krishnan, Punit~Singh Koura, Puxin Xu, Qing He, Qingxiao Dong, Ragavan Srinivasan, Raj Ganapathy, Ramon Calderer, Ricardo~Silveira Cabral, Robert Stojnic, Roberta Raileanu, Rohit Girdhar, Rohit Patel, Romain Sauvestre, Ronnie Polidoro, Roshan Sumbaly, Ross Taylor, Ruan Silva, Rui Hou, Rui Wang, Saghar Hosseini, Sahana Chennabasappa, Sanjay Singh, Sean Bell, Seohyun~Sonia Kim, Sergey Edunov, Shaoliang Nie, Sharan Narang, Sharath Raparthy, Sheng Shen, Shengye Wan, Shruti Bhosale, Shun Zhang, Simon Vandenhende, Soumya Batra, Spencer Whitman, Sten Sootla, Stephane Collot, Suchin Gururangan, Sydney Borodinsky, Tamar Herman, Tara Fowler, Tarek Sheasha, Thomas Georgiou, Thomas Scialom, Tobias Speckbacher, Todor Mihaylov, Tong Xiao, Ujjwal Karn, Vedanuj Goswami, Vibhor Gupta, Vignesh Ramanathan, Viktor Kerkez, Vincent Gonguet, Virginie Do, Vish Vogeti, Vladan Petrovic, Weiwei Chu,
  Wenhan Xiong, Wenyin Fu, Whitney Meers, Xavier Martinet, Xiaodong Wang, Xiaoqing~Ellen Tan, Xinfeng Xie, Xuchao Jia, Xuewei Wang, Yaelle Goldschlag, Yashesh Gaur, Yasmine Babaei, Yi~Wen, Yiwen Song, Yuchen Zhang, Yue Li, Yuning Mao, Zacharie~Delpierre Coudert, Zheng Yan, Zhengxing Chen, Zoe Papakipos, Aaditya Singh, Aaron Grattafiori, Abha Jain, Adam Kelsey, Adam Shajnfeld, Adithya Gangidi, Adolfo Victoria, Ahuva Goldstand, Ajay Menon, Ajay Sharma, Alex Boesenberg, Alex Vaughan, Alexei Baevski, Allie Feinstein, Amanda Kallet, Amit Sangani, Anam Yunus, Andrei Lupu, Andres Alvarado, Andrew Caples, Andrew Gu, Andrew Ho, Andrew Poulton, Andrew Ryan, Ankit Ramchandani, Annie Franco, Aparajita Saraf, Arkabandhu Chowdhury, Ashley Gabriel, Ashwin Bharambe, Assaf Eisenman, Azadeh Yazdan, Beau James, Ben Maurer, Benjamin Leonhardi, Bernie Huang, Beth Loyd, Beto~De Paola, Bhargavi Paranjape, Bing Liu, Bo~Wu, Boyu Ni, Braden Hancock, Bram Wasti, Brandon Spence, Brani Stojkovic, Brian Gamido, Britt Montalvo, Carl
  Parker, Carly Burton, Catalina Mejia, Changhan Wang, Changkyu Kim, Chao Zhou, Chester Hu, Ching-Hsiang Chu, Chris Cai, Chris Tindal, Christoph Feichtenhofer, Damon Civin, Dana Beaty, Daniel Kreymer, Daniel Li, Danny Wyatt, David Adkins, David Xu, Davide Testuggine, Delia David, Devi Parikh, Diana Liskovich, Didem Foss, Dingkang Wang, Duc Le, Dustin Holland, Edward Dowling, Eissa Jamil, Elaine Montgomery, Eleonora Presani, Emily Hahn, Emily Wood, Erik Brinkman, Esteban Arcaute, Evan Dunbar, Evan Smothers, Fei Sun, Felix Kreuk, Feng Tian, Firat Ozgenel, Francesco Caggioni, Francisco Guzmán, Frank Kanayet, Frank Seide, Gabriela~Medina Florez, Gabriella Schwarz, Gada Badeer, Georgia Swee, Gil Halpern, Govind Thattai, Grant Herman, Grigory Sizov, Guangyi, Zhang, Guna Lakshminarayanan, Hamid Shojanazeri, Han Zou, Hannah Wang, Hanwen Zha, Haroun Habeeb, Harrison Rudolph, Helen Suk, Henry Aspegren, Hunter Goldman, Ibrahim Damlaj, Igor Molybog, Igor Tufanov, Irina-Elena Veliche, Itai Gat, Jake Weissman, James
  Geboski, James Kohli, Japhet Asher, Jean-Baptiste Gaya, Jeff Marcus, Jeff Tang, Jennifer Chan, Jenny Zhen, Jeremy Reizenstein, Jeremy Teboul, Jessica Zhong, Jian Jin, Jingyi Yang, Joe Cummings, Jon Carvill, Jon Shepard, Jonathan McPhie, Jonathan Torres, Josh Ginsburg, Junjie Wang, Kai Wu, Kam~Hou U, Karan Saxena, Karthik Prasad, Kartikay Khandelwal, Katayoun Zand, Kathy Matosich, Kaushik Veeraraghavan, Kelly Michelena, Keqian Li, Kun Huang, Kunal Chawla, Kushal Lakhotia, Kyle Huang, Lailin Chen, Lakshya Garg, Lavender A, Leandro Silva, Lee Bell, Lei Zhang, Liangpeng Guo, Licheng Yu, Liron Moshkovich, Luca Wehrstedt, Madian Khabsa, Manav Avalani, Manish Bhatt, Maria Tsimpoukelli, Martynas Mankus, Matan Hasson, Matthew Lennie, Matthias Reso, Maxim Groshev, Maxim Naumov, Maya Lathi, Meghan Keneally, Michael~L. Seltzer, Michal Valko, Michelle Restrepo, Mihir Patel, Mik Vyatskov, Mikayel Samvelyan, Mike Clark, Mike Macey, Mike Wang, Miquel~Jubert Hermoso, Mo~Metanat, Mohammad Rastegari, Munish Bansal, Nandhini
  Santhanam, Natascha Parks, Natasha White, Navyata Bawa, Nayan Singhal, Nick Egebo, Nicolas Usunier, Nikolay~Pavlovich Laptev, Ning Dong, Ning Zhang, Norman Cheng, Oleg Chernoguz, Olivia Hart, Omkar Salpekar, Ozlem Kalinli, Parkin Kent, Parth Parekh, Paul Saab, Pavan Balaji, Pedro Rittner, Philip Bontrager, Pierre Roux, Piotr Dollar, Polina Zvyagina, Prashant Ratanchandani, Pritish Yuvraj, Qian Liang, Rachad Alao, Rachel Rodriguez, Rafi Ayub, Raghotham Murthy, Raghu Nayani, Rahul Mitra, Raymond Li, Rebekkah Hogan, Robin Battey, Rocky Wang, Rohan Maheswari, Russ Howes, Ruty Rinott, Sai~Jayesh Bondu, Samyak Datta, Sara Chugh, Sara Hunt, Sargun Dhillon, Sasha Sidorov, Satadru Pan, Saurabh Verma, Seiji Yamamoto, Sharadh Ramaswamy, Shaun Lindsay, Shaun Lindsay, Sheng Feng, Shenghao Lin, Shengxin~Cindy Zha, Shiva Shankar, Shuqiang Zhang, Shuqiang Zhang, Sinong Wang, Sneha Agarwal, Soji Sajuyigbe, Soumith Chintala, Stephanie Max, Stephen Chen, Steve Kehoe, Steve Satterfield, Sudarshan Govindaprasad, Sumit Gupta,
  Sungmin Cho, Sunny Virk, Suraj Subramanian, Sy~Choudhury, Sydney Goldman, Tal Remez, Tamar Glaser, Tamara Best, Thilo Kohler, Thomas Robinson, Tianhe Li, Tianjun Zhang, Tim Matthews, Timothy Chou, Tzook Shaked, Varun Vontimitta, Victoria Ajayi, Victoria Montanez, Vijai Mohan, Vinay~Satish Kumar, Vishal Mangla, Vítor Albiero, Vlad Ionescu, Vlad Poenaru, Vlad~Tiberiu Mihailescu, Vladimir Ivanov, Wei Li, Wenchen Wang, Wenwen Jiang, Wes Bouaziz, Will Constable, Xiaocheng Tang, Xiaofang Wang, Xiaojian Wu, Xiaolan Wang, Xide Xia, Xilun Wu, Xinbo Gao, Yanjun Chen, Ye~Hu, Ye~Jia, Ye~Qi, Yenda Li, Yilin Zhang, Ying Zhang, Yossi Adi, Youngjin Nam, Yu, Wang, Yuchen Hao, Yundi Qian, Yuzi He, Zach Rait, Zachary DeVito, Zef Rosnbrick, Zhaoduo Wen, Zhenyu Yang, and Zhiwei Zhao. 2024.
\newblock \href {http://arxiv.org/abs/2407.21783} {The llama 3 herd of models}.

\bibitem[{Elazar et~al.(2021)Elazar, Kassner, Ravfogel, Ravichander, Hovy, Schutze, and Goldberg}]{Elazar2021MeasuringAI}
Yanai Elazar, Nora Kassner, Shauli Ravfogel, Abhilasha Ravichander, Ed~Hovy, Hinrich Schutze, and Yoav Goldberg. 2021.
\newblock Measuring and improving consistency in pretrained language models.
\newblock \emph{ArXiv}, abs/2102.01017.

\bibitem[{Gao et~al.(2024)Gao, Xiong, Gao, Jia, Pan, Bi, Dai, Sun, Wang, and Wang}]{gao2024retrievalaugmentedgenerationlargelanguage}
Yunfan Gao, Yun Xiong, Xinyu Gao, Kangxiang Jia, Jinliu Pan, Yuxi Bi, Yi~Dai, Jiawei Sun, Meng Wang, and Haofen Wang. 2024.
\newblock \href {http://arxiv.org/abs/2312.10997} {Retrieval-augmented generation for large language models: A survey}.

\bibitem[{Gu et~al.(2023)Gu, Deng, and Su}]{gu-etal-2023-dont}
Yu~Gu, Xiang Deng, and Yu~Su. 2023.
\newblock \href {https://doi.org/10.18653/v1/2023.acl-long.270} {Don`t generate, discriminate: A proposal for grounding language models to real-world environments}.
\newblock In \emph{Proceedings of the 61st Annual Meeting of the Association for Computational Linguistics (Volume 1: Long Papers)}, pages 4928--4949, Toronto, Canada. Association for Computational Linguistics.

\bibitem[{He et~al.(2023)He, Cao, Chen, Liu, Li, Sun, and Zhao}]{he-etal-2023-lego}
Zhitao He, Pengfei Cao, Yubo Chen, Kang Liu, Ruopeng Li, Mengshu Sun, and Jun Zhao. 2023.
\newblock \href {https://doi.org/10.18653/v1/2023.findings-emnlp.613} {{LEGO}: A multi-agent collaborative framework with role-playing and iterative feedback for causality explanation generation}.
\newblock In \emph{Findings of the Association for Computational Linguistics: EMNLP 2023}, pages 9142--9163, Singapore. Association for Computational Linguistics.

\bibitem[{He et~al.(2024)He, Cao, Wang, Jin, Chen, Xu, Li, Jiang, Liu, and Zhao}]{he2024agentscourtbuildingjudicialdecisionmaking}
Zhitao He, Pengfei Cao, Chenhao Wang, Zhuoran Jin, Yubo Chen, Jiexin Xu, Huaijun Li, Xiaojian Jiang, Kang Liu, and Jun Zhao. 2024.
\newblock \href {http://arxiv.org/abs/2403.02959} {Agentscourt: Building judicial decision-making agents with court debate simulation and legal knowledge augmentation}.

\bibitem[{He et~al.(2025{\natexlab{a}})He, Liu, Li, Fung, Yan, Zhang, Huang, and Liu}]{he2025advancinglanguagemultiagentlearning}
Zhitao He, Zijun Liu, Peng Li, Yi~R Fung, Ming Yan, Ji~Zhang, Fei Huang, and Yang Liu. 2025{\natexlab{a}}.
\newblock \href {http://arxiv.org/abs/2502.14496} {Advancing language multi-agent learning with credit re-assignment for interactive environment generalization}.

\bibitem[{He et~al.(2025{\natexlab{b}})He, Lyu, Chen, Guo, and Fung}]{he2025matpbenchmllmgoodautomated}
Zhitao He, Zongwei Lyu, Dazhong Chen, Dadi Guo, and Yi~R. Fung. 2025{\natexlab{b}}.
\newblock \href {http://arxiv.org/abs/2506.06034} {Matp-bench: Can mllm be a good automated theorem prover for multimodal problems?}

\bibitem[{He et~al.(2025{\natexlab{c}})He, Polisetty, Fan, Huang, Wu, and Fung}]{he2025mmboundaryadvancingmllmknowledge}
Zhitao He, Sandeep Polisetty, Zhiyuan Fan, Yuchen Huang, Shujin Wu, and Yi~R. Fung. 2025{\natexlab{c}}.
\newblock \href {http://arxiv.org/abs/2505.23224} {Mmboundary: Advancing mllm knowledge boundary awareness through reasoning step confidence calibration}.

\bibitem[{Hendrycks et~al.(2021)Hendrycks, Burns, Basart, Zou, Mazeika, Song, and Steinhardt}]{hendryckstest2021}
Dan Hendrycks, Collin Burns, Steven Basart, Andy Zou, Mantas Mazeika, Dawn Song, and Jacob Steinhardt. 2021.
\newblock Measuring massive multitask language understanding.
\newblock \emph{Proceedings of the International Conference on Learning Representations (ICLR)}.

\bibitem[{Hu et~al.(2021)Hu, Shen, Wallis, Allen-Zhu, Li, Wang, Wang, and Chen}]{hu2021loralowrankadaptationlarge}
Edward~J. Hu, Yelong Shen, Phillip Wallis, Zeyuan Allen-Zhu, Yuanzhi Li, Shean Wang, Lu~Wang, and Weizhu Chen. 2021.
\newblock \href {http://arxiv.org/abs/2106.09685} {Lora: Low-rank adaptation of large language models}.

\bibitem[{Hu et~al.(2023)Hu, Luo, Wang, Cheng, Liu, and Sun}]{hu-etal-2023-wont}
Shengding Hu, Yifan Luo, Huadong Wang, Xingyi Cheng, Zhiyuan Liu, and Maosong Sun. 2023.
\newblock \href {https://doi.org/10.18653/v1/2023.acl-long.309} {Won`t get fooled again: Answering questions with false premises}.
\newblock In \emph{Proceedings of the 61st Annual Meeting of the Association for Computational Linguistics (Volume 1: Long Papers)}, pages 5626--5643, Toronto, Canada. Association for Computational Linguistics.

\bibitem[{Huang et~al.(2025)Huang, Fan, He, Polisetty, Li, and Fung}]{huang2025cultureclipempoweringclipcultural}
Yuchen Huang, Zhiyuan Fan, Zhitao He, Sandeep Polisetty, Wenyan Li, and Yi~R. Fung. 2025.
\newblock \href {http://arxiv.org/abs/2507.06210} {Cultureclip: Empowering clip with cultural awareness through synthetic images and contextualized captions}.

\bibitem[{Iyyer et~al.(2017)Iyyer, Yih, and Chang}]{iyyer-etal-2017-search}
Mohit Iyyer, Wen-tau Yih, and Ming-Wei Chang. 2017.
\newblock \href {https://doi.org/10.18653/v1/P17-1167} {Search-based neural structured learning for sequential question answering}.
\newblock In \emph{Proceedings of the 55th Annual Meeting of the Association for Computational Linguistics (Volume 1: Long Papers)}, pages 1821--1831, Vancouver, Canada. Association for Computational Linguistics.

\bibitem[{Jin et~al.(2024)Jin, Cao, Wang, He, Yuan, Li, Chen, Liu, and Zhao}]{jin2024rwku}
Zhuoran Jin, Pengfei Cao, Chenhao Wang, Zhitao He, Hongbang Yuan, Jiachun Li, Yubo Chen, Kang Liu, and Jun Zhao. 2024.
\newblock \href {http://arxiv.org/abs/2406.10890} {Rwku: Benchmarking real-world knowledge unlearning for large language models}.

\bibitem[{Li et~al.(2024)Li, Chen, Wang, Zhao, Liang, Hou, Liu, and Zhou}]{li2024mosaicitfreecompositionaldata}
Ming Li, Pei Chen, Chenguang Wang, Hongyu Zhao, Yijun Liang, Yupeng Hou, Fuxiao Liu, and Tianyi Zhou. 2024.
\newblock \href {http://arxiv.org/abs/2405.13326} {Mosaic-it: Free compositional data augmentation improves instruction tuning}.

\bibitem[{Liang et~al.(2024{\natexlab{a}})Liang, Wang, Bao, and Gao}]{liang-etal-2024-l}
Qiuyu Liang, Weihua Wang, Feilong Bao, and Guanglai Gao. 2024{\natexlab{a}}.
\newblock \href {https://aclanthology.org/2024.lrec-main.873/} {{L}{\textasciicircum}2{GC}:lorentzian linear graph convolutional networks for node classification}.
\newblock In \emph{Proceedings of the 2024 Joint International Conference on Computational Linguistics, Language Resources and Evaluation (LREC-COLING 2024)}, pages 9988--9998, Torino, Italia. ELRA and ICCL.

\bibitem[{Liang et~al.(2024{\natexlab{b}})Liang, Song, Wang, and Zhang}]{liang-etal-2024-learning}
Yuxin Liang, Zhuoyang Song, Hao Wang, and Jiaxing Zhang. 2024{\natexlab{b}}.
\newblock \href {https://doi.org/10.18653/v1/2024.knowledgenlp-1.4} {Learning to trust your feelings: Leveraging self-awareness in {LLM}s for hallucination mitigation}.
\newblock In \emph{Proceedings of the 3rd Workshop on Knowledge Augmented Methods for NLP}, pages 44--58, Bangkok, Thailand. Association for Computational Linguistics.

\bibitem[{Lin et~al.(2024)Lin, Diesendruck, Du, and Abraham}]{lin2024batchpromptaccomplish}
Jianzhe Lin, Maurice Diesendruck, Liang Du, and Robin Abraham. 2024.
\newblock \href {http://arxiv.org/abs/2309.00384} {Batchprompt: Accomplish more with less}.

\bibitem[{Liu et~al.(2023)Liu, Chen, Sakai, and Wu}]{Liu2023AFL}
Qijiong Liu, Nuo Chen, Tetsuya Sakai, and Xiao-Ming Wu. 2023.
\newblock \href {https://api.semanticscholar.org/CorpusID:263891105} {A first look at llm-powered generative news recommendation}.
\newblock \emph{ArXiv}, abs/2305.06566.

\bibitem[{Mangrulkar et~al.(2022)Mangrulkar, Gugger, Debut, Belkada, Paul, and Bossan}]{peft}
Sourab Mangrulkar, Sylvain Gugger, Lysandre Debut, Younes Belkada, Sayak Paul, and Benjamin Bossan. 2022.
\newblock Peft: State-of-the-art parameter-efficient fine-tuning methods.
\newblock \url{https://github.com/huggingface/peft}.

\bibitem[{Maynez et~al.(2020)Maynez, Narayan, Bohnet, and McDonald}]{maynez-etal-2020-faithfulness}
Joshua Maynez, Shashi Narayan, Bernd Bohnet, and Ryan McDonald. 2020.
\newblock \href {https://doi.org/10.18653/v1/2020.acl-main.173} {On faithfulness and factuality in abstractive summarization}.
\newblock In \emph{Proceedings of the 58th Annual Meeting of the Association for Computational Linguistics}, pages 1906--1919, Online. Association for Computational Linguistics.

\bibitem[{Peng et~al.(2023)Peng, Galley, He, Cheng, Xie, Hu, Huang, Liden, Yu, Chen et~al.}]{peng2023check}
Baolin Peng, Michel Galley, Pengcheng He, Hao Cheng, Yujia Xie, Yu~Hu, Qiuyuan Huang, Lars Liden, Zhou Yu, Weizhu Chen, et~al. 2023.
\newblock Check your facts and try again: Improving large language models with external knowledge and automated feedback.
\newblock \emph{arXiv preprint arXiv:2302.12813}.

\bibitem[{Qin et~al.(2025)Qin, Dong, Zhang, Dong, Huang, Yang, Khademi, Zhang, Awadalla, Fung, Chen, Cheng, and Wei}]{qin2025scalinglawssyntheticdata}
Zeyu Qin, Qingxiu Dong, Xingxing Zhang, Li~Dong, Xiaolong Huang, Ziyi Yang, Mahmoud Khademi, Dongdong Zhang, Hany~Hassan Awadalla, Yi~R. Fung, Weizhu Chen, Minhao Cheng, and Furu Wei. 2025.
\newblock \href {http://arxiv.org/abs/2503.19551} {Scaling laws of synthetic data for language models}.

\bibitem[{Reddy et~al.(2019)Reddy, Chen, and Manning}]{reddy-etal-2019-coqa}
Siva Reddy, Danqi Chen, and Christopher~D. Manning. 2019.
\newblock \href {https://doi.org/10.1162/tacl_a_00266} {{C}o{QA}: A conversational question answering challenge}.
\newblock \emph{Transactions of the Association for Computational Linguistics}, 7:249--266.

\bibitem[{Ren et~al.(2023)Ren, Zhao, Liu, Wu, Wen, and Wang}]{ren2023tometwostageapproachmodelbased}
Ruiyang Ren, Wayne~Xin Zhao, Jing Liu, Hua Wu, Ji-Rong Wen, and Haifeng Wang. 2023.
\newblock \href {http://arxiv.org/abs/2305.11161} {Tome: A two-stage approach for model-based retrieval}.

\bibitem[{Son et~al.(2024)Son, Baek, Nam, Jeong, and Kim}]{son2024multitaskinferencelargelanguage}
Guijin Son, Sangwon Baek, Sangdae Nam, Ilgyun Jeong, and Seungone Kim. 2024.
\newblock \href {http://arxiv.org/abs/2402.11597} {Multi-task inference: Can large language models follow multiple instructions at once?}

\bibitem[{Sun et~al.(2023)Sun, Shen, Zhou, Zhang, Chen, Cox, Yang, and Gan}]{sun2023principledrivenselfalignmentlanguagemodels}
Zhiqing Sun, Yikang Shen, Qinhong Zhou, Hongxin Zhang, Zhenfang Chen, David Cox, Yiming Yang, and Chuang Gan. 2023.
\newblock \href {http://arxiv.org/abs/2305.03047} {Principle-driven self-alignment of language models from scratch with minimal human supervision}.

\bibitem[{Wang et~al.(2024)Wang, Kodner, and Rambow}]{wang2024exploringzeroshotcapabilitiesllms}
Zhengxiang Wang, Jordan Kodner, and Owen Rambow. 2024.
\newblock \href {http://arxiv.org/abs/2406.10786} {Exploring the zero-shot capabilities of llms handling multiple problems at once}.

\bibitem[{Yang et~al.(2024)Yang, Yang, Hui, Zheng, Yu, Zhou, Li, Li, Liu, Huang, Dong, Wei, Lin, Tang, Wang, Yang, Tu, Zhang, Ma, Yang, Xu, Zhou, Bai, He, Lin, Dang, Lu, Chen, Yang, Li, Xue, Ni, Zhang, Wang, Peng, Men, Gao, Lin, Wang, Bai, Tan, Zhu, Li, Liu, Ge, Deng, Zhou, Ren, Zhang, Wei, Ren, Liu, Fan, Yao, Zhang, Wan, Chu, Liu, Cui, Zhang, Guo, and Fan}]{yang2024qwen2technicalreport}
An~Yang, Baosong Yang, Binyuan Hui, Bo~Zheng, Bowen Yu, Chang Zhou, Chengpeng Li, Chengyuan Li, Dayiheng Liu, Fei Huang, Guanting Dong, Haoran Wei, Huan Lin, Jialong Tang, Jialin Wang, Jian Yang, Jianhong Tu, Jianwei Zhang, Jianxin Ma, Jianxin Yang, Jin Xu, Jingren Zhou, Jinze Bai, Jinzheng He, Junyang Lin, Kai Dang, Keming Lu, Keqin Chen, Kexin Yang, Mei Li, Mingfeng Xue, Na~Ni, Pei Zhang, Peng Wang, Ru~Peng, Rui Men, Ruize Gao, Runji Lin, Shijie Wang, Shuai Bai, Sinan Tan, Tianhang Zhu, Tianhao Li, Tianyu Liu, Wenbin Ge, Xiaodong Deng, Xiaohuan Zhou, Xingzhang Ren, Xinyu Zhang, Xipin Wei, Xuancheng Ren, Xuejing Liu, Yang Fan, Yang Yao, Yichang Zhang, Yu~Wan, Yunfei Chu, Yuqiong Liu, Zeyu Cui, Zhenru Zhang, Zhifang Guo, and Zhihao Fan. 2024.
\newblock \href {http://arxiv.org/abs/2407.10671} {Qwen2 technical report}.

\bibitem[{Zhang et~al.(2024)Zhang, Diao, Lin, Fung, Lian, Wang, Chen, Ji, and Zhang}]{zhang-etal-2024-r}
Hanning Zhang, Shizhe Diao, Yong Lin, Yi~Fung, Qing Lian, Xingyao Wang, Yangyi Chen, Heng Ji, and Tong Zhang. 2024.
\newblock \href {https://doi.org/10.18653/v1/2024.naacl-long.394} {{R}-tuning: Instructing large language models to say {\textquoteleft}{I} don`t know'}.
\newblock In \emph{Proceedings of the 2024 Conference of the North American Chapter of the Association for Computational Linguistics: Human Language Technologies (Volume 1: Long Papers)}, pages 7113--7139, Mexico City, Mexico. Association for Computational Linguistics.

\end{thebibliography}
